# Hybrid Intelligent System

## *A Self-Adaptive Network Protection System*

*"Using Soft Computing and Open-Source Tools"*

**Author**

**Mohamed Hassan**

# Table of Contents

# List of Figures

# List of Tables

# Acronyms

**IDS:** *Intrusion Detection System*
**IPS:** *Intrusion Prevention System*
**SSH:** *Secure SHell*
**IP:** *Internet Protocol*
**TCP:** *Transport Control Protocol*
**UDP:** *User Datagram Protocol*
**ICMP:** *Internet Control Message Protocol*
**AI:** *Artificial Intelligence*
**EC:** *Evolutionary Computation*
**EA:** *Evolutionary Algorithm*
**GA:** *Genetic Algorithm*
**FL:** *Fuzzy Logic*
**BLP:** *Bell-LaPadula Model*
**R2L:** *Remote to Local*
**U2R:** *User to Root*
**MPI:** *Messages Passing Interface*
**PVM:** *Parallel Virtual Machine*
**API:** *Application Programming Interface*
**UTM:** *Unified Threat Management*
**AIS:** *Artificial Immune System*
**DSS:** *Decision Support System*
**LAN:** *Local Area Network*
**WAN:** *Wide Area Network*
**OS:** *Operating System*
**LDB:** *Learning Database*

# Abstract

The aim of this study is to conduct a profound research in biological-inspired computing, fuzzy logic, and linguistic control methods (rules based systems) under the context of soft computing in order to build a self-adaptive, and intelligent network threats assessment and protection system. But how can we build such as intelligent system without complexity? the idea behind this research is to find a simple approach to synthesis a self-adaptive intelligent network protection sensor without getting in deep packet analysis and/or into complex pattern elicitation process. Nowadays, the immense growth of the global network (Internet) and pervasive usage of e-commerce, networks expanding significantly in businesses, users, social, and communities, which indeed influencing our lives; computers, smart phones, tablets, and virtual stores became increasingly available almost everywhere. Any computer security concern or network protection system in particular aim to control, who allowed to pass through the network and who is not, simply is a kind of process control. Information environment and its security concerns have become the most important issue in any environment using computers and networks, people responsible for IT in businesses trying to follow and conduct the necessary steps (procedures) to protect their networks and/or IT infrastructure either from inside or outside. With that vast increase in computers and global networks, the Internet and networks security threats have emerged, accordingly the hacking term came to light.

In this research project, simplicity was an aim to synthesise theories, methods, and tools to build self-organized intelligent network protection system with focus on combination of firewall and Intrusion Detection/Prevention mechanism, barely few research focused on such combination. But why yet another network security system? did today networks including the global grid (Internet) needs more protection systems? what will be different? perhaps, these questions jump to the mind, which necessarily needs an answers.

Trying to design/develop flexible and adaptive security oriented approaches is a challenge, therefore in this study an endeavour was made to design a novel approach for network threats protection system.

# CHAPTER 1

# Introduction

Almost all computing, from its nature sense, including cluster based environment (cloud), super computing machines, and any system that performs computations are considered as *hard computing*. They are using mathematical approaches to problem solving, and they inherit their basic characteristics from mathematics (centred to use numbers and symbols). On the opposite side *soft computing* considered to be a tools or methods, which are inspired by nature and particularly biology, they are using biological approaches to problem solving, where mathematics does not play a major role as it does in engineering problem solving methods. (Rajasekaran and Pai, 2003)

Research in artificial intelligence (AI) directed to study the human mental capabilities, including reasoning, understanding, imagination, recognition, perception, prediction and ultimately emotions. (Hopgood, 2001)

But to build such a machine and improving our understanding of intelligence, a new methods is needed to achieve most of the above and getting toward a computational theory of perceptions, the current methods is not sufficient, because when solutions to problems are cross-disciplinary (in their nature), soft computing promises to be a powerful method for finding solutions to problems faster than the conventional methods, yet accurately and acceptably. (Rajasekaran and Pai, 2003)

## 1.1 Background

In this project we shall present a synthesis of hybrid network automated (self-adaptive) security threats discovery and prevention system, by using unconventional techniques and methods, including fuzzy logic and biological-inspired algorithms under the term of soft computing methods. The tools to achieve the project objectives (section 1.5) will be based on open-source (free software), which gives more freedom to view/alter any existing code and build on it. To grasp the necessary information about computing adaptation we need to understand what previous researches or studies have discovered, hence it is essentials to highlight about thoughts, theories, and results of the others in the same discipline (chapter 2 for more details).

The idea behind self-adaptive systems and how machines can improve their behaviour by learn (experiencing) from their environment, which they can change the behaviours or actions, commonly based on accumulating experience.

Another thing that should be considered when talking about self-adaptive system, the reaction of the self-organized system, which should be intelligent, accordingly terms such as self-adaptive and intelligent system should be clear, what make machine/program intelligent?; how it could be self-organized? and how machine could learn or think? which it seems to be a philosophical question, however answering these questions was aim of many researchers for decades, also it would be necessary trying to answer it by regarding to this research.

Most of the current network security management systems, such as firewall, IDS, and IPS acting as barrier between local network and one or more external networks to compel network traffic for a certain security policy by deciding which packets to let pass through and which to deny, on the set of rules defined by the network administrator. Any error in defining the rules may compromise the system security by letting undesired traffic pass through or blocking the desired traffic. Manually defined rules often results conflicting, redundant or overshadowed rules, which creates anomalies in the firewall policy. (Chaure and Shandilya, 2010)

Conventional intrusion detection and prevention techniques, such as firewalls, access control or encryption, have failed to fully protect networks and systems from increasingly sophisticated attacks and malwares. As a result, intrusion detection systems (IDS) have become an indispensable component of security infrastructure to detect these threats before they inflict widespread damage. (Wu and Banzhaf, 2010)

Many research efforts have been focused on how to effectively and accurately construct detection models. Starting with a combination of expert systems and statistical approaches to acquiring knowledge of normal or abnormal behaviour had turned from manual to automatic. Artificial intelligence and machine learning techniques were used to discover the underlying models from a set of training data. Commonly used methods were rule based induction, classification and data clustering.(Wu and Banzhaf, 2010) In chapter 2 we will focus in more details on pertinent literature review.

## 1.2 Research Questions

Research and investigation in science and all aspect around us are usually start with a question, i.e. why sky is blue? what is the origin of species? and so on, in this research an endeavour was made to answer the following study questions, consequently to achieve the research objectives.

### 1.2.1  In nutshell

- What is computing intelligence?

- Why soft computing?

- How soft computing methods can be integrated to build such intelligent system?

- What system needs to be built and what will protect?

- Why this system should be a self-adaptive?

- How can build knowledge (which could be extracted from simple data) from grid activity behaviours? (further research)

- Can we build rational machine, which think like human? (further research)

## 1.3  Problem Statement

Nowadays, Internet has immensely pervasive usage, people are using it is nearly about %35 percent of the world population[1] and definitely increasing; consequently computer networks have expanded significantly in use and numbers. This expansion makes them more vulnerable to attack by malicious agents. Many current intrusion detection systems (IDS) are unable to identify unknown or mutated attack modes. (Fries, 2008)

In the conventional signature-based intrusion detection systems (in its nature) the mode of operation is passive and storage-limited, even the if the IDS in the inline mode (prevention status), there still need of signature storage. Their operation depends upon catching an instance of an intrusion or virus and encoding it into a signature that is stored in its anomaly database, providing a window of vulnerability to computer systems during this time. Further, the maximum size of an Internet Protocol-based message requires the database to be huge in order to maintain possible signature combinations. (Haag et al., 2007)

Security threats is not only prevented by IDS or IPS there are many more techniques and solution trying to protect the operated network, starting from low level network security, through the systems to the applications. Another example one of the current technology, firewalls is to be considered as one of the major security part in any networked environment, but it has a lack of packet taxonomy capabilities, which (most of the current firewalls) are using net-filter based approach, just to filter the packet upon the header on a static manner (as pre-configured rules

---

[1]http://www.internetworldstats.com/stats.htm

based) by checking source and destination IP's and port numbers and some other header flags. Firewall cannot profile the network traffic for anomaly behaviours, it considered to be a barrier between networks that allow or deny traffic based on particular settings, even (few enterprise firewall types) the payload based inspection firewalls lack of profiling and/or has no self-adaptation mechanism, also needs user management/administration capabilities.[2] [3]

Another problem with traditional rules-match based firewalls and many other similar tools, which require a manual configuration and settings, which is may not be comprehended by novice users or even network system administrators, which they will need a comprehensive training to deal with rules complexity (especially in the large environment). Therefore, it comes the need of self-adaptive/self-learning techniques based on network activity or behaviour to tackle the given problems.

Table 1.1: Common Open-Source security management systems

| IDS | Firewall | Anti-Virus/Spam |
|---|---|---|
| **Host-Based** | **Application Level** | **Host** |
| OSSEC | Netfilter application layer patch | ClamAV |
| Nagios | AppArmor | |
| **Network-Based** | **Kernel Level** | **Network** |
| Snort, Bro, and Suricata | Netfilter/Iptables | SpamAssassin & dSpam |

## 1.4  Research Focus

In the past few years plethora of research and development leads to a new technology, which has discovered in every field and particularly in computer science, hence it will be a complex task to present all of such broad research area (IT/IS threats and soft computing), therefore the aim is to narrow this research project by trying to achieve the research objectives (hereinafter) precisely and make things simple as much as we can. Of-course in this research we will not attempt to build an i-Robot machine, even the greatest scientist couldn't, it is a matter of contribution in the AI field from another perspective.

Therefore, this project approach is meant to emphasis on studying soft computing methods, including fuzzy logic and evolutionary algorithms to build a self-adaptive, intelligent network threat prevention system using open source software and the mentioned methods.

---

[2] https://www.paloaltonetworks.com/products/features/application-visibility.html
[3] http://www.juniper.net/us/en/products-services/security/netscreen/ns5400/

## 1.5 Study Aim & Objectives

**The aim of this study** can be formed from the previous discourse in other word from the research focus itself, as this study aim is to build an intelligent, self-adaptive network threat protection sensor, by using soft computing methods and open-source tools, which this system can use a new methods and algorithms.

### 1.5.1 Research Statement

**Research statement** To study soft computing methods, including fuzzy logic and evolutionary algorithms to build a self-adaptive, intelligent network threat protection system using open source tools combined with all or part of soft-computing methods.

### 1.5.2 Study Objectives

This project divided into three major objectives, so according to the problem statement (section 1.3), the first part emphasize on research for finding new methods to tackle current obstacles in network threats management systems, followed by the second part focus on developing a state of the art (new generation) intelligent threats protection system, word protection in the last phrase to emphasise that the proposed system is not a management system, hence it is a self-adaptive intelligent system; the later part focus on how to evaluate the idea.

### 1.5.3 Research

This considered to be the theoretical part of the project, the main objective is to study soft computing methods, including fuzzy logic and evolutionary algorithms. In addition and as part of fuzzy logic, explore linguistic based system (IF-THEN rules).

### 1.5.4 Development

Develop a state-of-the-art (next generation) self-adaptive, and intelligent network threats protection system, in the current state a software based system will be presented.

### 1.5.5 Evaluation

Evaluate results against project tasks, which will be encapsulated in evaluation matrix, and demonstrate experimental testing phase as well.

### 1.5.6 Encapsulate objectives into tasks

- Study fuzzy logic (imprecision and uncertainty decision making perspective).
- Investigate in the nature-based computing area, biological inspired algorithms in particular.
- Study linguistic based system (rules/classifier based system).
- Study network threat management system (focus on DOS attack).
- Build an intelligent self-adaptive system.
- Demonstrate the proposed idea in a software based prototype (Validation).

## 1.6 Value of this Research

Network security threats management becomes a crucial issue when designing any information based environment, while the conventional methods that currently used have some obstacles as stated in problem statement, so in order to improve network security threats capabilities there is a need of integration of a new methods to tackle the current deficiency, trying to contribute in this field with a new synthesis of soft computing methods to add on with efficient way.

### 1.6.1 What is Different in this Research?

Yet another network threats management system or IDS/IPS, many researchers have demonstrate different approaches and techniques to achieve network security goal, which protect networks either from outside or inside threats, nevertheless why more researches still interest in the same subject, simply because networks/Internet become vital in social life, businesses, industries, scientific research, governmental environment, and even day-to-day lifestyle.

Therefore, a question may leaping to mind, what makes this project different?, dichotomy of the answer: Synthesis of the research, and Intelligent adaptation. This is not a management system, which means there is no human (administrators) interaction with the system, it is a self-organized intelligent system. Implies more than protection system (most of current research focus only on IDS), firewall and IDS with inline mode (IPS), and can adapt more information security techniques (further research).

To tackle the current network security approaches, a combination of two methods (Fuzzy Logic & Evolutionary Algorithms) have been applied in this research, so it is a Hybrid system.

**The new approaches in this research are:**

- Combine firewall and IDS/IPS techniques (All-in-One network threats protection system). *Almost all current researches focused only either IDS (most) or firewall (few)*

- Add Anti-virus/Anti-spam/Anti-maleware to the above combination. *(Further Research)*.

- Combine two methods (fuzzy logic and evolutionary algorithms) to detect network threats (Hybrid).

- Using the evolutionary algorithms to build simple intelligent, and self adaptive system (Intelligent).

- The proposed system design has the scalability feature and could be distributed. *(Further Research)*.

- Not-related network threats (Host threats) could be added to this research. *(Further Research)*

- In short, an All-in-One intelligent, not manageable (self-adaptive) information security threats protection system.

The following sections will provide a short overview in order to answers some of the research questions (section 1.2), in addition a need of clarification of what the proposed system will protect?, hence a highlight of information security risks in general and network threats in particular will be considered.
**Questions will try to answer it:**

- What is intelligent system?

- What is self-adaptation in computing?

- What is the common information environment threats?

- What system needs to be built and what will protect?

## 1.7 Meaning of Intelligent System

In this research, the proposed system is a combination of tools, methods, algorithms, programs, and of-course machine (hardware), integrate all or part of them on a unified simple intelligent system, it is a kind of artificial intelligence approach. Therefore a brief explanation will be useful to understand what the meaning of intelligent in this research, of-course is not a profound elucidation of the subject, since AI is a cross-disciplinary field. According to (Brownlee, 2011), (Konar, 2000) AI is a systems that think and act like humans or think and act rationally (Russell and Norvig, 2003), but can really computer systems think/act like human? do all human have intelligence? and what about emotion, evolution, and cognition?; AI is a seductive field of research and many researchers trying to attain the AI main goal, which is build a human-like system . Computers was built based on the use of mathematical and logical approaches to solve problems, accordingly an endeavour was made to develop and enhance computers, but still needs logical, sequential input feeding to electronics circuits to process it in mathematical and logical manner, ultimately producing output. What is human intelligence? Is it problem solving (but you have to think about the problem nature if it easy or complex), is it reasoning ability, gain knowledge faster than others, thinking speed, actually there is no unique definition of human intelligence. But when it comes to machine or program intelligence a need of evolving and learning should be adapted, therefore to find a way to evolve systems/programs to change dynamically within its environment was an aim for many researcher; accordingly this research is a try to combine some techniques to reach intelligence and adaptation approach in one system.

One of the most significant work in this field presented by Koza, John (1992), which summarize the idea in his early book "Genetic Programming" One impediment to getting computers to solve problems without being explicitly programmed is that existing methods of machine learning, artificial intelligence, self-improving systems, self-organizing systems, neural networks, and induction do not seek solutions in the form of computer programs. Instead, existing paradigms involve specialized structures which are nothing like computer programs (e.g., weight vectors for neural networks, decision trees, formal grammars, frames, conceptual clusters, coefficients for polynomials, production rules, chromosome strings in the conventional genetic algorithm, and concept sets). Each of these specialized structures can facilitate the solution of certain problems, and many of them facilitate mathematical analysis that might not otherwise be possible. However, these specialized structures are an unnatural and constraining way of getting computers to solve problems without being explicitly programmed. Human programmers do not regard these specialized structures as having the flexibility necessary for programming computers, as evidenced by the fact that computers are not commonly programmed in the language of weight vectors, decision trees, formal grammars, frames, schemata, conceptual clusters, polynomial coefficients, production rules, chromosome strings, or concept sets. (Koza, 1992)

Evolutionary computation has proven techniques as a optimization technique, and automation design system for its unparalleled mechanism, because it inherited from nature, biological nature in particular, which indeed unique.

Existing network security technology and/or firmware based device (appliance), including firewalls, IDS[4], IPS[5], UTM[6] have many remarkable capabilities, and there still more contribution in the same conventional manner, even the smart devices have some lacks of deduction, autonomy, and intelligent capabilities. The obstacle of deduction from perception-based information.

As Zadeh L. A. (Zadeh, 2006)p.2 proposed *"As a basic example, assume that the question is q: What is the average height os Swedes?, and the available information is p: Most adult Swedes are tall. Another example is: Usually Robert returns from work at about 6 p.m. What is the probability that Robert is at home at 6:15 pm? Neither bivalent logic nor probability theory provide effective tools for dealing with problems of this type."*

The way that conventional methods deal with the current matching rule-based mechanism in network security either in firewalls or in IDS's is not sufficient as described in problem statement section (1.3). Therefore, an endeavour was by many researcher (see chapter 2) to find a new and different methods to contribute in this field, by using soft computing techniques, fuzzy logic and biological inspired methods to enhance and tackle the current obstacle in network threat management system.

Soft computing is not just a random mixture of disciplines such fuzzy logic, neural nets, and evolutionary algorithms, but a discipline in which each constituent contributes a distinct methodology for addressing problems in its domain, in a complementary rather than competitive way. (Tettamanzi and Tomassini, 1998)

## 1.8 Meaning of Self-Adaptive System

Self-adaptation is a set of process performed continually throughout the lives of all living things, it is an abstract process of establishing learning through classification, in other words adaptation is a respond to environmental changes as in biological evolution. (Patterson, 1990) A self-adaptive system is a set of interacting process through sensing new conditions and adapt with new changes accordingly. [7]

### 1.8.1 How systems can be self-organized?

By leaning living things can gain new knowledge form past experience learn new knowledge an existing one should be presented, thus we classify the new concept by relate (cluster) it with something we already know, if we cannot build this cluster of knowledge based on the previous one; then we must create a new structure for understanding for this new knowledge or pattern. (Jones, 2006)
**The meaning for self-adaptation in this research** is to evolving network packets behaviour as a classifier by using evolutionary computation (genetic/cell algorithms), this evolution can be adapted by find learning mechanism to create a program/system that can be able to learn and classify knowledge (pattern) using accumulated experience.

---

[4]IDS: Intrusion Detection System

[5]IPS: Intrusion Prevention System

[6]UTM: Unified Threat Management

[7]http://en.wikipedia.org/wiki/Adaptive_system

Accordingly, to build a self-adaptive intelligent system, two things that system should do:

1. **The ability to choose and then to act.**

2. **The ability to experience things (learning), and be able to adapt and evolve.**

## 1.9 Information Environment Security and Threats

### 1.9.1 What is the Information Environment?

The information environment is a man-made construct based on the idea that the existence and proliferation of information and information systems has created a new operating environment. This environment can be used by organizations, individuals, to gain an advantage over their opponents. As such, the information environment can be used to gain knowledge and make decisions. (Romanych, 2007)

### 1.9.2 Information Environment Model

The model consists of three interrelated dimensions; the physical, informational, and cognitive:
**The physical dimension** → is that part of the information environment which coexists with the physical environments of air, land, sea, and space. It is where information and communication systems and networks reside, whether they are either technology or human-based.
**The cognitive dimension** → is the individual and collective consciousness that exists in the minds of human beings. It is where perceptions are formed, and more importantly, where decisions are made.
**The informational dimension** → is an abstract, non-physical space created by the interaction of the physical and cognitive domains. As such, it links the reality of the physical dimension to the human consciousness of the cognitive dimension. It is the means through which individuals and organizations communicate. (Romanych, 2007)

### 1.9.3 Bell-LaPadula (BLP) Information Security Model

Bell–LaPadula Model (BLP) is a data protection module was developed by David Elliott Bell and Leonard J. LaPadula [8], based on data and information multilevel security policy, which mean each data have a label (tag) describing its security level range from the most sensitive (e.g."Top Secret"), down to the least sensitive (e.g., "Unclassified" or "Public"). [8]

- Simple Security Rule: Subjects cannot read data from a higher level than they are cleared for. For example, a Secret clearance holder cannot read Top Secret data.

- *-Property (or Star Property): Subjects cannot write any data below their own levels (often called no-write-down). Similar to the simple security rule, a user is not allowed to write to a lower level.

- Strong Star Property: Subjects can only read and write data within their classification level. [9]

### 1.9.4 Generic Information Security Model

- **Host Security:** Tools that involved in system security (stand alone server/machine), such as Host Intrusion Detection System (HIDS). It is a kind of auditing system integrity; most common tools → Tripwire and OSSEC.

- **Network Security:** Tools that involved in networks protection, such as unified threat management (UTM), which include: → Firewalls, Intrusion Detection System (IDS's), Proxies, VPN's, and TCP/IP network management and analysis.

- **Communication Security:** Ultimately, the communication security is the matter of securing your communication medium, in other words it is the matter of cryptography, such as → Secure Socket Layer (SSL), Transport Layer Security (TLS), Secure Shell (SSH), Pretty Good Privacy (PGP), and Gnu Privacy Guard (GPG), or any other tools that may use cryptography to cipher the communications between two nodes.

---

[8] http://en.wikipedia.org/wiki/Bell%E2%80%93LaPadula_model
[9] http://www.infosecschool.com/bell-lapadula-model/

In addition to the above, it is substantial to monitor and audit systems, one of the most popular tools is syslog/rsyslog or syslog-ng. Typically, syslog-ng is used to manage log messages and implement centralized logging, where the aim is to collect the log messages of several devices on a single, central log server.[10] But why logging is vital to any systems and/or networks, simply because you can track the activity of any suspicious network or system behaviours, which in our case tracking any hacking attempt.

### 1.9.5 IT Security Goals

IT security means keep your information under your full direct control, preventing access to it by anyone else without your permission, and be aware of the dangers posed to allow someone access to your private information. The security goals of any IT/IS infrastructure consisting of three elements, Confidentiality, Integrity, and Availability. (Behrouz, 2008)

**Confidentiality**

Confidentiality is the most common aspect of information security. While using the computers and the local/global networks we aim to protect our confidential information. An organization needs to guard against those malicious actions that endanger the confidentiality of its information. In other words, it is a concealment of sensitive information, which is a major concern. (Behrouz, 2008)

**Integrity**

Information needs to be changed constantly, so integrity means that changes need to be done only be authorized entities and through authorized procedures. Integrity violation is not necessarily to be the result of a malicious act, so it could be from any External element, such as power surge, which may create unwanted changes in some information. (Behrouz, 2008)

**Availability**

The information created and stored by an organization needs to be available to authorized entities all the time, the unavailability of information is as harmful for an organization as the lack of confidentiality or integrity. (Behrouz, 2008)

---

[10]Extracted from syslog-ng (Linux) manual page

## 1.10 Network Threats

Network treats considered to be an attempt of someone or a party that exploits a weakness in a software or a system (vulnerability) over the network, which could be used by the attacker to compromise the system without authorisation.

### 1.10.1 Most common types of IT attacks

The types of attack that any IT infrastructure could face, consists of two parts *Active & Passive* attacks, in *active* attack the aim of intruder is to harm the system and/or modify/alter/delete the data, while in *passive* attack (a kind of eavesdropping) the attacker aim is to sniff the communication and may get a copy of it, for later use. **Some Examples:**

**Passive attack (Confidentiality)**

Sniffing: $\rightarrow$ Monitor network traffic.

**Active attack (Integrity)**

Spoofing: $\rightarrow$ Attacker impersonate somebody else (Masquerading).
Repudiation: $\rightarrow$ Denying of identity.
Modification: $\rightarrow$ Altering data.

**Active attack (Availability)**

Denial-of-Service (DoS): $\rightarrow$ Flooding the network with massive random packets that target host cannot handle.

**Another perspective in IT threats**

In addition to the above, also threats can be categorised as follow:

- **Remote to User Attacks (R2L):** A remote to user attack is an attack in which a user sends packets to a machine over the internet, which s/he does not have access in order to expose the machines vulnerabilities and exploit privileges which a local user would have on the computer e.g. xlock, guest, xnsnoop, phf, sendmail dictionary etc.

- **User to Root Attacks (U2R):** These attacks are exploitations in which the hacker starts off on the system with a normal user account and attempts to abuse vulnerabilities in the system in order to gain super user privileges e.g. perl, xterm.

- **Probing:** Probing is an attack in which the hacker scans a machine or a networking device in order to determine weaknesses or vulnerabilities that may later be exploited so as to compromise the system. This technique is commonly used in data mining e.g. saint, portsweep, mscan, nmap etc.

### 1.10.2 Common TCP/IP Treats/Attack

- TCP Segment Format
- TCP Disconnection
- IP Address Spoofing
- IP Fragment Attacks
- TCP Flags (Research focus)
- Syn Flood (Research focus)
- Ping of Death (Research focus)
- UDP Flood Attack
- Connection Hijacking
- ARP, DNS, E-Mail Spoofing

## 1.11 Taxonomy of Protection Systems

Nowadays, the risk of information environment security flow has been immensely raised, therefore a need of effective IT defence solutions are need to protect information environment from incidents, many organizations have implemented security threat management solutions such as anti-virus software, anti-spam systems, firewalls, and intrusion detection systems (IDSs). In this section an overview of security management system will be propounded

### 1.11.1 Network-based Protection System

**Firewall**

Firewall is a kind of middle ground between networks either internal (LAN) or external (WAN/Internet), the main role of the firewall is to isolate the local (internal) LAN from the Internet or those that are connected to it (external). Firewall could be a software among other services running on a server or a device (appliance) that only act as a firewall, which placed between organization's LAN and WAN in order to provide a simple way to control the amount and kinds of traffic that will pass between the networks. (Garfinkel and Spafford, 1996)

The main function of firewall is to control/restrict network traffic between networks, in other words you must define what kinds of data pass and what kinds are blocked this called a firewall policy, by allowing or denying data passed through your network an access control mechanism should be created, which called a firewall rules. (Garfinkel and Spafford, 1996)

**Uses of Firewalls**

Firewalls used as part of any organization network protection strategy among others layers of information security. There are some obvious threats from external networks, hence a firewall in the middle between local network and external one is mandatory.

Table 1.2: Firewalls usage

| Category | Use of the Firewall |
| --- | --- |
| Data Access | Can be used to block traffic to particular networks and/or websites |
| Monitor Traffic | Can be used to monitor communication between LAN & WAN |
| Secure Communication | Can be used to encrypt network traffic (VPN) |
| Connection Limitation | Can be used to control traffic between networks |
| Acceleration | Can be used as caching proxy and web content filter |

**Network-based Intrusion Detection System (NIDS)**

One of the major objectives in network security is to detect incidents, minimizing exposure, and assessing and understanding the risk on all levels, not only an exercise in building perimeter defences. (Zalewski, 2005) On a basic level, network intrusion detection is the process of determining when unauthorized people are attempting to break into your network. (Cox and Gerg, 2004) The main purpose of network intrusion detection system (NIDS) to detect unauthorised network based access, such as attempts to login to your system, access unprotected network shares, and flooding network by sending a carefully crafted sequence of packets to a network server and ultimately crashing it (Denial of service attack), DOS considered to be an intrusion, also considered a successful attack because it occupies resources that would have been employed somewhere else.

Probing someone network with port scans or ping (ICMP) perhaps not an intrusion, but it is a sign that may soon start doing something more hostile. Network probing activity is also considered intrusion, and expect network protection system to warn whenever things such as these happen. (Cox and Gerg, 2004)

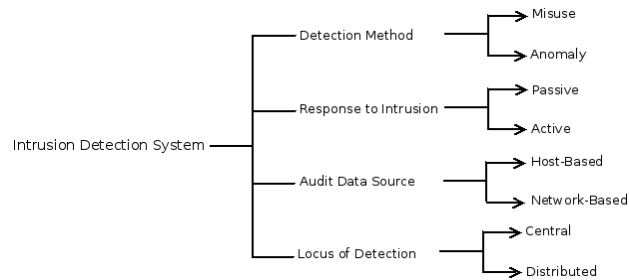

Figure 1.1: Characteristics of IDS in general

Figure source: (Wu and Banzhaf, 2010)

There are two generally accepted categories of intrusion detection techniques: misuse detection (also called signature based) and anomaly detection (packet behaviour based). Misuse detection refers to techniques that characterize known methods to penetrate a system. These penetrations are characterized as a pattern or a signature that the IDS looks for. The pattern/signature might be a static string or a set sequence of actions. System responses are based on identified penetrations. In anomaly detection, the system administrator defines the baseline, or normal, state of the networks traffic load, breakdown, protocol, and typical packet size. The anomaly detector monitors network segments to compare their state to the normal baseline and look for anomalies.[11]

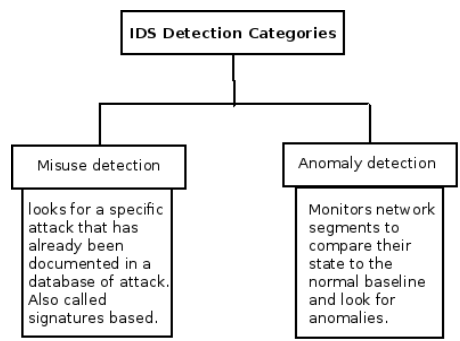

Figure 1.2: IDS Detection Categories

**Intrusion Detection Traffic Analysis compared to Content Analysis**

The strength of a traffic analysis approach is that it is possible to examine and record every packet that passes by on the network, it is a kind of network moni-toring. The weakness of this approach is that some attacks can only be detected by content analysis. Another approach to intrusion detection is to examine the contents of packets for certain strings that indicate an attack, this is a content, or string based approach. The strength of the content based approach is that it is able to collect all of the data of a connection once an attack string is detected. The weakness of this approach is that it is virtually impossible to analyse and record all of the traffic on a network, therefore many services, or ports are filtered out making the system blind to attacks using these ports or services.[12]

---
[11]http://www.webopedia.com/TERM/I/intrusion_detection_system.html

**Firewall Compared to IDS (short overview)**

To highlight the differences between Firewall and IDS, which both considered to be a network security techniques: Firewall looks outwardly for intrusions in order to stop them from happening, and limit access between networks to prevent intrusion and do not signal an attack from inside the network. On the other hand an IDS evaluates a suspected intrusion once it has taken place and signals an alarm. An IDS also watches for attacks that originate from within a system. This is traditionally achieved by examining network communications, identifying heuristics and patterns (often known as signatures) of common computer attacks, and taking action to alert operators. A system that terminates connections is called an intrusion prevention system, and is another form of an application layer firewall.[13]

### 1.11.2 Host-based Protection System

As of the network based IDS is to detect (normally) the incidents from external world (outside the local network), in some case local network should be aware of inside intruder (caused by viruses). On the other hand host-based IDS is a software to be installed on individual hosts such as anti-virus, anti-spam mechanism, personal firewall, filesystem integrity checker (tripwire), access control lists (ACL's), and mandatory access control (MAC) which enforce security policies either on the kernel space level or user level (SELinux). Many host-based protection systems are hybrid mixing all or some of these solutions (anti-virus, personal firewall, etc...). Host-based protection technologies is out for scope of this project, we will focus on network based protection mechanisms, a complete security protection system will undergo further research, such as hybrid system (host-network protection system) emphasises on adaptation in any environment with intelligent protection mechanism (further research).

---

[12] http://grox.net/doc/security/stepbystep.htm
[13] http://en.wikipedia.org/wiki/Intrusion_detection_system

### 1.11.3   What the proposed system will protect?

This is an obvious question, what this proposed prototype will protect, since many of information environment threats (especially network based threats) have been identified for long time now and categorized as follow:

Table 1.3: Categories of IT threats

| | |
|---:|:---:|
| Malicious | Cause loss, embarrassment or inconvenience for spite or commercial advantage. |
| Mischievous | Cause damage or inconvenience by exposing system vulnerabilities. |
| Fraudulent | Intended to access to privileged information. |
| Consequential | Loss, effect, exposure or damage as a consequence of omission or other activity. |
| Failure | Failure or loss of a system or connection (Availability). |

Source: DrayTek online white paper [14]

Therefore this research focus on network availability threats, Denial of Service (DoS) attack in particular, an endeavour was made in this research to achieve project objectives (section 1.5.2) with the focus of DoS attack as one of the common network threats.

### 1.11.4   A short overview of Denial of Service Attack (DoS)

DoS attack categorized as intruder that attacker attempt to prevent legitimate users of a service from using or having access to that service. The idea behind a DoS attack is to takes up so much of a shared resource that none of the resource is left for other users, DoS attacks compromise the availability, which is resulting a degradation or loss of service. Resources could be categorize in two main group (A) Shared resource on a local host such as a overload system process, disk attack by fill a disk partition with unwanted data, etc.. Group (B) Network DoS attacks, which the most common attacks used nowadays. The aim of the network DoS attack is to deny/loss web or network/internet service, by flooding the target with so much packets requests that it cannot respond (the server) to it, which will lead loss/degradation of the services or information. (Garfinkel and Spafford, 1996)

There is another evolving from DoS attack called Distributed Denial of Service Attack (DDoS), which has evolved over time. The difference between DoS and DDoS attacks is, in DDoS attacker could overwhelm the resources of victims with floods of packets through multiple and distributed computers in different geographical locations. (Dhanjani, 2003)

---

[14]http://www.draytek.co.uk/products/network_threats.html

### 1.11.5   Approaches in Nutshell

Synopsis of our endeavour to achieve the research project objectives as follow:

- Studying fuzzy logic in how to make decisions on imprecise and uncertain situations.

- Learning and studying biological-inspired based algorithms.

- Synthesize/Integrate tools that needed to build the mentioned intelligent system/sensor.

- Plan & Designing a prototype theoretically (How it works).

- Implement the prototype.

## 1.12   Hypothesis

The idea behind this research is to design a simple self-organized intelligent network threats protection system without complexity, a straightforward and an easy to use system (adaptive). Simple by integrating the current open source tools such as netfilter Linux kernel modules, tcpdump, snort, bro (IDS's), and functionality of Linux syslog program and/or custom script(s) to generate particular logs (database). Further, an endeavour was made to simplifying evolutionary algorithm and fuzzy logic rules to automatically feed the decision engine (system brain) with updated and best rules that match the current network traffic behaviour. The research philosophy is to imitate biological adaptation by simulating the nature evolution of living animals and plants, as one of intelligence characteristics is to choose and then act, the main aim of this research project is to build an efficient intelligent network protection system, which can select random solution (rules) and be pro-active, as result dropping suspicious network activity. In this research an assumption was made that some static network service/program/rules (i.e IDS, firewall) could be changed according to the current environment changes, therefore there is a need of new methods that can be evolved during the time based on experience (learning), in the next few papers more explanation will be presented.

## 1.13   Dissertation Structure

This project divided into seven chapters starting with chapter 1, where addressing the research questions and project aim/objectives. The following chapter (chapter 2) focus on relevant research studies that previously done (Literature Review) and its limitations. Chapter 3 discusses the research methodology and working environment. In chapter 4 a brief overview of soft-computing methods and its characteristics. Chapter 5 discusses the proposed system architecture and how the system work, followed be chapter 6 system prototype and prove of concept. Ultimately, chapter 7 the findings and conclusion. Appendix A, presented how to install and configure service and tools that may used during this research project, in appendix B, describe how to use this tools; in appendix C, all codes have been used to demonstrate the proposed system, ultimately appendix D, shows proposed system screenshots.

# CHAPTER 2
# Review of Pertinent Literature

As describe in chapter one (section 1.9) and to articulate the differences between this research and the others work, a short reminder of the research objectives should be beneficial, in which we can map each objective with the recent research study. The aim of this research is to build an intelligent, self-adaptive network threat protection sensor, by using soft computing methods and open-source tools, such system can use a new methods and algorithms,

**Summery of Project Objectives**

- Study fuzzy logic (imprecision and uncertainty decision making perspective).

- Investigate in evolutionary computing discipline, genetic algorithms in particular.

- Study linguistic based system (rules/classifier based system).

- Study network threat management system (focus on DOS attack).

- Build an intelligent self-adaptive system.

- Demonstrate the proposed idea in a software based prototype.

Many endeavours was made in computer security field and network protection approaches in particular, therefore the need of studying this relevant work is substantial, so a taxonomy of this work with focusing on network threats protection systems in needed. Most of previous pertinent work focusing on network security was considered either artificial immune system (AIS) or IDS/NIDS with focusing on usage of one of soft computing tools such as fuzzy logic or evolutionary algorithms, only few papers represented hybrid system (includes fuzz logic as classifier and GA) (Panda et al., 2012), and (Amza et al., 2011).

To identify the characteristic of this research, an initial study should be conducted by narrowing this research approaches as shown in below section. Soft computing, including fuzzy logic and neural network and evolutionary algorithms, considered to be a subsidiary from artificial intelligence, which has broader scope and encompasses more disciplines to facilitate complex problems solving, but cannot be considered as soft computing, because AI approaches relay on conventional way of reasoning such as crisp logic, symbolic reasoning and computational language processing among others methods, which imbibe their roots from mathematics. (Tettamanzi and Tomassini, 1998)

According to Zeleznikow & Nolan (2001)a decision support system (DSS). is a computer-based information system that combines models and data in an attempt to solve non-structured problems with extensive user involvement.(Zeleznikow and Nolan, 2001)

Gouda & Liu (2007) proposed in their paper, a new firewall mechanism called "Structured Firewall" which a proposed designs of firewall using decision diagram instead of a sequence of often conflicting rules. And they also proposed program to converts the firewall decision diagram into a compact, yet functionally equivalent, sequence of rules. This method addresses the consistency problem because a firewall decision diagram is conflict-free. It addresses the completeness problem because the syntactic requirements of a firewall decision diagram force the designer to consider all types of traffic. It also addresses the compactness problem because in the second step we use two algorithms (namely FDD reduction and FDD marking) to combine rules together, and one algorithm (namely firewall compaction) to remove redundant rules. (Gouda and Liu, 2007)

Also Wu Xiaonan & Banzhaf Wolfgang (2010) have stated the obstacles and insufficiency in the conventional network security mechanisms they said: *"Traditional intrusion prevention techniques, such as firewalls, access control or encryption, have failed to fully protect networks and systems from increasingly sophisticated attacks and malwares. As a result, intrusion detection systems (IDS) have become an indispensable component of security infrastructure to detect these threats before they inflict widespread damage."*. (Wu and Banzhaf, 2010)

Since the firewalls and intrusion detection systems in its nature is a rule/signature based systems, which they needs a regular update for their knowledge base or rules structure. Intrusions are usually polymorph, and evolve continuously, it is difficult to detect new intrusions, thus new approach should be applied to evolve the IDS. Misuse detection will fail easily when facing unknown intrusions, hence they need such as update, as well as the firewalls needs rules update in case of new policy applied.

## 2.1 Network-based Protection System

Mostaque Md.(Md. and Hassan, 2013) presents a few research papers regarding the foundations of intrusion detection systems, the methodologies and good fuzzy classifiers using genetic algorithm which are the focus of current development ef-forts and the solution of the problem of Intrusion Detection System to offer a real- world view of intrusion detection. Ultimately, a discussion of the upcoming technologies and various methodologies which promise to improve the capability of computer systems to detect intrusions is offered. He used KDD Cup 1999 "Computer Network Intrusion Detection" competition data set to detect novel attacks by unauthorized users in network traffic. He consider an attack to be novel if the vulnerability is unknown to the target's owner or administrator, even if the attack is generally known and patches and detection tests are available. He used fuzzy log method/rules and genetic algorithm to identify the normal and abnormal behaviour computer networks, and fuzzy inference logic can be applied over such rules to determine when an intrusion is in progress. The main problem with this process is to make good fuzzy classifiers to detect intrusions.

### 2.1.1 Firewall

In (Chaure and Shandilya, 2010) paper, authors made an endeavour to highlight current firewall techniques to covers the advancements of various approaches proposed by researchers in this field. Their paper also discusses the policies for creating, modifying and sustaining the rule sets of firewall in such a way that makes the rule set optimal and free from known anomalies. Since manual processing for detecting anomalies in firewall is complex and often error-prone. Any minor change in the rule set of firewall leads to the requirement of rigorous analysis for maintaining the consistency and efficiency of firewall mechanism.

### 2.1.2 Network Intrusion Detection System (NIDS)

Catania and Garino (Catania and Garino, 2012) published a good paper of how automate network based IDS, they survey the most relevant works in the field of automatic network intrusion detection. In their analysis they considers several features required for truly deploying each one of the reviewed approaches. This wider perspective can help us to identify the possible causes behind the lack of acceptance of novel techniques by network security experts.

**NIDS using Evolutionary Algorithms**

Ren Hui Gong(Gong et al., 2005) and others, proposed GA-based intrusion detection approach contains two modules applied in two stages. First the training stage, where a set of classification rules are generated from network audit data using the GA in an offline environment. The second stage is the intrusion detection, where the generated rules are used to classify incoming network connections in the real-time environment. Once the rules are generated, the intrusion detection is simple and efficient. In the following sections, we focus our discussions on deriving the set of rules using GA.

Statistical method of intrusion detection attempts to predict future events based on what (events) have already occurred. Researchers found new approaches to compare the recent behaviour of a computer user of a with previous behaviour and any significant deviation is considered as intrusion. This approach requires construction of a pattern for normal user behaviour. Predictive pattern generation uses a rule base of user profiles defined as statistically weighted event sequences. (Abraham et al., 2007) and (Dhak and Lade, 2012)

In the same paper (Abraham et al., 2007) authors attempts to illustrate how genetic programming techniques could be deployed for detecting known types of attacks. They described three genetic programming techniques.

### 1. Linear Genetic Programming (LGP)

In linear GP programs are linear sequences of instructions, such as *{instruction 1, 2, ..., instruction n}*. The number of instructions can be fixed, meaning that every program in the population has the same length, or variable, meaning that different individuals can be of different sizes.(Poli et al., 2008)
The main characteristics is basic unit of evolution of this method is a native machine code instruction that runs on the floating-point processor unit (FPU). LGP uses a specific linear representation of computer programs. Instead of expressions of a functional programming language (like LISP ), so programs of an imperative language (like C) can be evolved.(Abraham et al., 2007)

### 2. Multi Expression Programming (MEP)

In this method genetic chromosome generally encodes a single expression (computer program). By contrast, a Multi Expression Programming (MEP) chromosome encodes several expressions. The best of the encoded solution is chosen to represent the chromosome by supplying the fitness of the individual. MEP genes are represented by sub-strings of a variable length, and each number of genes per chromosome is constant. This number defines the length of the chromosome, so each gene encodes a terminal or a function symbol. A gene that encodes a function includes pointers towards the function arguments. Function a guments always have indices of lower values than the position of the function itself in the chromosome.

### 3. Gene Expression Programming

Gene expression programming, a genotype/phenotype genetic algorithm (linear and ramified), is presented here for the first time as a new technique for the creation of computer programs. Gene expression programming uses character linear chromosomes composed of genes structurally organized in a head and a tail. The chromosomes function as a genome and are subjected to modification by means of mutation, transposition, root transposition, gene transposition, gene recombina-tion, and one- and two-point recombination. The chromosomes encode expression trees which are the object of selection.(Ferreira, 2001)

Wei Li(Li, 2004) emphasis on his implementation of both temporal and spatial information of network connections in encoding the network connection information into rules in IDS. One network connection and its related behaviour can be translated to represent a rule to judge whether or not a real-time connection is considered an intrusion. These rules can be modeled as chromosomes inside the population. The population evolves until the evaluation criteria are met.

Also, A.A. Ojugo(Ojugo et al., 2012) presents a genetic algorithm based approach, which employs a set of classification rule derived from network audit data and the support-confidence framework, utilized as fitness function to judge the quality of each rule. The software implementation is aimed at improving system security in networked settings allowing for confidentiality, integrity and availability of system resources.

**Fuzzy Logic Approaches in NIDS**

Patrick LaRoche and others proposed in (LaRoche et al., 2009) a first step to build a fuzzing system that can learn to adapt for vulnerability analysis. By developing a system that learns the packets that are required to be transmitted towards targets, using feedback from an external network source, they makes a step towards having a system that can intelligently explore the capabilities of a given security system. In order to validate our system's capabilities we evolve a variety of port scan patterns while running the packets through an IDS, with the goal to minimizes the alarms raised during the scanning process. LaRoche stated in the above paper that fuzzing can be used at the transport level, while nodes communicating.

Jonatan Gomez and Dipankar Dasgupta(Gomez and Dasgupta, 2002) proposes a technique to generate fuzzy classifiers using genetic algorithms that can detect anomalies and some specific intrusions. The main idea is to evolve two rules, one for the normal class and other for the abnormal class using a profile data set (a preprocessed DARPA data set is used) with information related to the computer network during the normal behaviour and during intrusive (abnormal) behaviour. A collection of fuzzy sets, called fuzzy space, defines the fuzzy linguistic values or fuzzy-classes that an object can belong to. With fuzzy spaces, fuzzy logic allows an object to belong to different classes at the same time. This concept is helpful when the difference between classes is no well defined. It is the case in the intrusion detection task, where the difference between the normal and abnormal class are not well defined. With these linguistic concepts, atomic and complex fuzzy logic expressions can be built. For example (as they propose in the same paper):

*IF condition THEN consequent [weight] Where,*

- *condition is a complex fuzzy expression, i.e., that uses fuzzy logic operators and atomic fuzzy expressions*

- *consequent is an atomic expression, and*

- *weight is a real number that defines the confidence of the rule.*

*IF x is HIGH and y is LOW THEN*
*pattern is normal, otherwise intrusion.*

### 2.1.3 Artificial Immune Systems

Simon T. Powers (Powers and He, 2008) proposed a hybrid artificial immune system with the aim of combining the advantages of IDS anomaly detection and misuse detection. They used Kohonen Self Organising Map method[1] to flag and categorised network connections that have anomalous behaviour, allowing higher-level information, in the form of cluster membership, to be extracted.

Jungwon Kim and others (Kim et al., 2007) review the current approaches in artificial immune system (AIS), they introduce suitable intrusion detection problems to AIS researchers. They have presented the requirements for an effective network-based IDS. These requirements can be applied not only to a network-based IDS, but to any type of IDS. These requirements are of particular interest because they could be fulfilled by mechanisms inspired by features of the human immune system. The Human Immune System (HIS) can detect and defend against harmful and previously unseen invaders, so can a similar system be built for our computers? Perhaps, those systems would then have the same beneficial properties as the HIS such as error tolerance, adaptation and self-monitoring.

#### Hybrid Approach

Terrence P. Fries, one of few researchers combine fuzzy logic with evolutionary algorithms as he proposed in (Fries, 2010) an evolutionary fuzzy rule-based intrusion detection system, as he presents a fuzzy inference for IDS with evolutionary optimization which overcomes both poor anomaly detection rate and high number of false positives, by developing a set fuzzy rules which identify intruders. The algorithm for constructing the IDS is comprised of two distinction parts. First, a genetic algorithm is used to establish an optimal subset of the communication features which are necessary to identify intrusions. Second, a set of fuzzy rules is optimized using another genetic algorithm. He proposed the use of GA in the first

---

[1]http://www.ai-junkie.com/ann/som/som1.html

stage to build a training set of rules, as of encoding GA chromosomes in binary string of 0's and 1's with the number of bits equal to the total number of features. Each bit represents a particular feature. If the bit is a '1', it indicates the attribute is to be used for training, while a '0' indicates the attribute is not to be used. A GA determines the optimal set of features to be used for training the rule set. Then he used fuzzy logic as classifier system that uses if-then rules, by mapping features weight with fuzzy linguistic variables and the consequent classification is qualified with a certainty factor.(Fries, 2010)

What Fries proposed is slightly similar to the idea of this project, but the difference is (as the aim of this project and for further research) to find way to combine detection with action in real-time mode in simple way.

## 2.2  Conclusion on Literature Review

After studying many related papers that are closed to this research ideas, some issues have been found, most of these papers are complex, particularly in applied chosen method(s), such as Gene Expression Programming and Multi Expression Programming, which both are complex genetic techniques, even the idea behind them is great. Also some results was not efficient compared to the traditional IDS/Firewall methods (false positive and negative alarms rate), in addition there was a lack on the combination of multi security mechanisms for network protection such IDS combined with firewall and HIDS; another thing regarding the knowledge base (learning mechanisms) used for the IDS intelligence, which have to be in more cellular/cluster way, such devices (IDS/firewall) communication via standard protocol and/or agent based program.

Most of the papers that proposed related ideas are using same data set for (DARPA - KDDCUP 99) classify different types of attack connections (training and testing data), this data set was collected back 1998 and 1999, which already outdated for the new types of attack. Attackers are becoming rapidly more sophisticated, far outpacing IDS; so simply you can define the known types of attack and make pre-installed (static) rules and then increment on-top of this rule-base your own learning data set. Another interesting thing, that DARPA tcpdump raw data is usually used in offline mode, therefore most of the experiments were done offline, not in real-time network activity, which is vital point to consider.

The aim of this study is efficiency and simplicity, it will required further research and improvement, but the idea behind it (section 1.12) is to make an All-in-One intelligent system able to communication with other devices/systems (cell-based cluster environment) to exchange the knowledge (learning data), to tackle the unknown (zero-day) attacks.

Whilst, simplicity is an aim for this research, many ideas from this literature review were valuable and very effective in the design, it is a small contribution on what others researchers did, yet another add-ons for a hope of achieving research objectives efficiently.

# CHAPTER 3

# Soft Computing

Soft computing term encompasses fuzzy logic, evolutionary computation and its algorithms, and neural network, which is focus on new methods to solve problems. In this chapter a brief overview will be presented on fuzzy logic and biological inspired computation (neural networks out of scope of this research), this overview is necessary to grasp a concept about fuzzy logic and evolutionary computing to be applied in the proposed system.

## 3.1 Boolean Logic & Fuzzy Logic

Fuzzy set theory was propounded by Lotfi Zadah in 1965 (Zadeh, 1965), since then many of theoretical developments have taken a leading role in advancing this field. Fuzzy logic derived from fuzzy set theory as boolean logic which had its roots from the classical set theory (crisp logic).

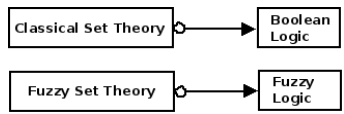

Figure 3.1: Roots of boolean logic and fuzzy logic

Boolean logic is a mathematical theory of classical logic and probabilities based on binary system which has only two numbers 0 and 1 and it also called propositional logic which propositions may be "true" or "false". Fuzzy logic also provides a mathematical framework for representing uncertainty in the sense of imprecision and partial truth, it is a degree values between 0 and 1 (i.e » $0 \to 0.25, 0.5, 0.75, ... \to 1$). (Tettamanzi and Tomassini, 1998)

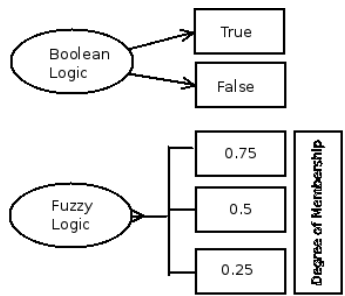

Figure 3.2: Boolean/Crisp logic and Fuzzy logic

**Classical logic Membership Function**

Crisp set: Indicator function of element X of set A is defined by: (Rajasekaran and Pai, 2003)

$$\mu_A(\chi) = \begin{cases} 0 & \text{if}, \chi \notin A, \\ 1 & \text{if}, \chi \in A. \end{cases}$$

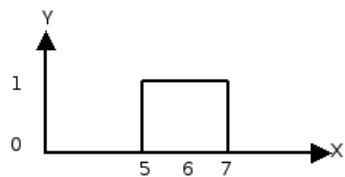

Figure 3.3: Crisp sets indicator (membership) function

**Fuzzy logic Membership Function**

Fuzzy set: Membership function in fuzzy logic encompasses elements that satisfy imprecise properties of membership which defined by: (Rajasekaran and Pai, 2003)

$$\mu_A(\chi) = \in [0,1]$$

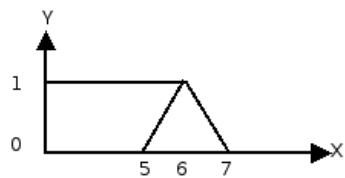

Figure 3.4: Fuzzy sets membership function

### 3.1.1 Crisp (Classical/Boolean) Logic

Before over viewing the classical logic, as brief introduction for traditional set theory should be presented, in the following section a definition of sets and its characteristic will be presented.

**Concept of sets**

Sets theory was propounded by Georg Cantor in 19th-century, a set is a collection of objects (elements) with the same type or different type within the same context, but not discretionary (must be precise):
*Examples:*

- All students in university. ⇒ Is a set.

- All students are more than 20 years of old. ⇒ Is a set.

- N = {1,2,3,...} ⇒ Is a set of natural numbers.

**But**

- Colours are inclined red. ⇒ Is not a set, because of imprecision and uncertainty.

- Tall players in a football team. ⇒ Is not a set, because height not determined.

In crisp set theory a universe of discourse (universal set) is a set of objects (elements) having the same characteristics and reference to a particular context (that why called universe of discourse); universal set is denoted by $E$. (Rajasekaran and Pai, 2003)

**Representation of sets**

Set is usually denoted by capital letters A,B,C,..., and its elements are denoted by small letters a,b,c,...,.

$$\text{Given a set of } A \text{ whose elements are } a_1, a_2, a_3, \ldots, a_n,$$

Ways to represent a set:

**1. Descriptive Form:**

Describe set elements in words

$$A = \textit{All students in a university.}$$

**2. Tabulation or List Method**

Listing all elements in a set separated by commas within a curly brackets {}

$$A = \{a_1, a_2, a_3, \ldots, a_n\}$$

$a_1, a_2, a_3, \ldots, a_n$ are called *members or elements* of the set.

**3. Rule Method**

This method is based on common properties shared by all set members.

$$R = \{\chi : 0 < \chi < 1\}$$

Real numbers between 0 and 1. This example cannot be represented by tabulation (listing) method.[1]

### 3.1.2 Fuzzy Logic

Fuzzy logic to be considered a superset of Boolean (traditional) logic that has been extended to handle the concept of partial truth, the truth values between 0 and 1 or extremely " true" and extremely "false", so it is a kind of degree of set membership. Fuzzy logic is a continuum grade of membership (Zadeh, 1965), also it could be considered as degree based logic ranging between zero and one 0→1 and in between (e.g 0.3, 0.5, 0.7) Fuzzy Logic is a kind of logic using graded or qualified statements rather than ones that are strictly true or false. The results of fuzzy reasoning are not as definite as those derived by strict logic, but they cover a larger field of discourse. (Zadeh, 1984)

To grasp the basic understanding of fuzzy set, let $U$ be a universe of discourse and its objects (elements) denoted by $x$ which is a generic element of $U$.

A fuzzy set of $A$ in $U$ is defined by membership function $\mu_A(\chi)$ also called the membership degree in which $x$ belongs to $A$, associating to every element in $U$ a real number in between [0,1].

Table 3.1: Crisp sets and Fuzzy sets in comparison

| Crisp Sets | Fuzzy Sets |
|---|---|
| Elements either belong to the set or not. | Elements can partially be in the set. |
| Membership function has only two values, either 0 or 1. | Membership function has degree in between 0 and 1. |
| No other values allowed between 0 and 1. | Other values between 0 and 1 are allowed. |

---

[1] http://www.emathzone.com/tutorials/algebra/definition-and-representation-of-set.html

**Fuzzy Propositions**

As in crisp logic, fuzzy propositions is a statement which have uncertain (fuzzy) truth value. Thus given $\tilde{P}$ to be a fuzzy proposition, $T(\tilde{P})$ represents the truth value $(0 \to 1)$ which attached to $\tilde{P}$.

The fuzzy membership value associated with the fuzzy set $\tilde{A}$ for $\tilde{P}$ is same as the fuzzy truth value $T(\tilde{P})$. (Rajasekaran and Pai, 2003)

$$T(\tilde{P}) = \mu_{\tilde{A}}(\chi) \; where \; 0 \leq \mu_{\tilde{A}}(\chi) \leq 1$$

*Example:*

$$\tilde{P} : \text{Socrates is honest.}$$
$$T(\tilde{P}) = 0.7, \text{ if } \tilde{P} \text{ is partly true.}$$
$$T(\tilde{P}) = 1, \text{ if } \tilde{P} \text{ is absolutely true.}$$

$T(\tilde{P}) = Truth$ value (in degrees) of $(\tilde{P}) \to [0,1]$

Also, fuzzy logic similar to traditional/crisp logic that supports connectives:

- Negation : $\neg$
- Conjunction (AND) : $\wedge$
- Disjunction (OR) : $\vee$
- Implication (IF-THEN) : $\Rightarrow$

Table 3.2: Fuzzy logic connectives

| Symbol | Connective | Usage | Definition |
|---|---|---|---|
| $\neg$ | Negation | $\neg \tilde{P}$ | $1-T(\tilde{P})$ |
| $\wedge$ | Conjunction | $\tilde{P} \wedge \tilde{Q}$ | min $(T(\tilde{P}), T(\tilde{Q}))$ |
| $\vee$ | Disjunction | $\tilde{P} \vee \tilde{Q}$ | max $(T(\tilde{P}), T(\tilde{Q}))$ |
| $\Rightarrow$ | Implication | $\tilde{P} \Rightarrow \tilde{Q}$ | $\neg \tilde{P} \vee \tilde{Q} =$ max $(1-T(\tilde{P}), T(\tilde{Q}))$ |

**Fuzzy Quantifiers**

Fuzzy quantifiers are numbers used to quantify logical predicates, as in crisp logic fuzzy logic propositions are quantified by fuzzy quantifiers, which categorised in two classes:

1. Absolute quantifiers $\to \Re$ (Real numbers).

2. Relative quantifiers → interval [0, 1].

Table 3.3: Fuzzy logic quantifiers

| Absolute Quantifier | Relative Quantifier |
|---|---|
| about *10* | almost all |
| much greater than *100* | about half |
| some where around *5* | most |

**Example (Quantified propositions):**

There are about 10 packets in incoming network traffic, where intrusions score are high.

$\tilde{P}$ *: There are Q Z's in $\tilde{A}$* where *Z* is fuzzy set that defined as follow:

$$\mu_z(i) = \mu_F(V(i)) \; \forall \; i \in \tilde{A}$$

$\tilde{Q}$ = (about 10) → Absolute quantifier.

$i$ = packets (is an individual from a given set $\tilde{A}$).

$\tilde{A}$ = incoming traffic.

$V(i)$ = intrusions score from incoming packets (is a variable associated to the individual from set $\tilde{A}$ that takes values from a universe $\tilde{E}$).

$F$ = Fuzzy set that represents *high* degree of attack score.

**Fuzzy Inference**

Fuzzy inference is a computational procedures used for evaluating linguistic descriptions, also referred to approximate reasoning. There are two major fuzzy inferring procedures:

- Generalized Modus Ponens (GMP)
- Generalized Modus Tollens (GMT)

$$\text{IF } antecedent \text{ THEN } consequent$$

$$\text{IF } \tilde{P}1 \; (\chi_1,\ldots,\chi_n) \text{ THEN } \tilde{Q}1 \; (y_1,\ldots,y_m)$$

where *$\tilde{P}1$* and *$\tilde{Q}1$* represent fuzzy predicates on independent and dependent variables. (Tettamanzi and Tomassini, 1998)

Such above examples is important in classification and rules based systems which have been used by many researcher in both misuse and anomaly network intrusion detection. (Stach et al., 2005), (Gomez and Dasgupta, 2002), (Liangxun et al., 2011)

## 3.2 Biological Inspired Computation

The main characteristic of nature is *evolution*, and biology considered to be a nature-science which concerned with the study of living things and their vital processes.[2] Inspiration from biology and its nature of evolution has led to several successful algorithmic approaches. Such methods are frequently used to tackle complex problems (especially optimisation problem). Evolutionary algorithms, particle swarm optimisation, and ant colony have its inspiration from nature and biological life such algorithms have found numerous applications for solving problems from computational biology, engineering, logistics, network and artificial intelligence. Bio-inspired algorithms have achieved tremendous success when applied to many computation problems in recent years. (Neumann and Witt, 2010)

## 3.3 The Art of Evolutionary Computing

Biological cell that contain DNA, chromosomes, genes, and proteins are evolution products. Evolutionary computing is a new kind of classical algorithm that emulates the biological evolutionary process in intelligent search, machine learning and optimization problems. (Konar, 2000)

Popular evolutionary algorithms include the Genetic Algorithm, Genetic programming, Evolutionary programming, and practical swarm optimization, all are inspired from nature and it main characteristic (Evolution). The evolutionary process is considered to be an adaptive process and is typically applied to search and optimization domains. (Brownlee, 2011)

**Genetic Algorithms** is one of the most prevalent algorithms, recently it has been used as alternative to conventional methods in computer security in general and in intrusion detection in particular; it is considered to be a stochastic search algorithm. Genetic algorithms make it possible to explore a far greater range of potential solutions to a problem than do conventional programs. (Holland, 1992)

In order to make genetic algorithm successful, the algorithm should involve the following three steps, which makes it very easy to apply such methods to a newly given problem.

---

[2]http://en.wikipedia.org/wiki/Biology

1. Choose a representation of possible solutions.

2. Determine a function to evaluate the quality of a solution.

3. Define operators that produce from a current set of solutions a new set of solutions.

(Neumann and Witt, 2010)

## 3.4 Why use Bio-inspired Algorithms?

- Solve difficult problems.
- Extensible.
- Hybridize.
- Less programming code.
- Help to create adaptive rule-bases.
- Efficient in machine learning.

## 3.5 Why using Soft-Computing in this research?

Soft-computing encompasses fuzzy logic, neural networks, and biological (evolutionary) inspired computation. The conventional computer science approaches that could model and precisely analyse only relatively simple systems, contrariwise it is a framework to deal with uncertainty and imprecision (Fuzzy Logic) and automate/create a working computer program to randomly search for optimal solution for a given problem (Evolutionary Algorithms). In this research an endeavours was made to find new methods to prevent computer network threats (e.g intruder), intruders is dynamic and changing rapidly and current protection methods cannot overcome evolved threats; which indeed running in uncertain environment (network traffic behaviours). There are a need for new tools/methods to achieves the goal of automatic programming by genetically make off-spring of new population of computer programs using the principles of Darwinian natural selection and biologically inspired operations; and fuzzy logic as tool for decision making to manipulate imprecise and noisy data. In addition fuzzy logic could create sets that in degree values between 0 and 1, to represent rules/situations/events are not well defined. In this case linguistic (if-then-rules) based approach can applied, by taking minimum of set of events (0) or maximum (1) and in between (i.e., 0.5, 0.75, 0.95) and generate rules as follow:

$$\textit{IF events} >= \textit{0.75, THEN high (intrusion} \rightarrow \textit{DROP)}$$
$$\textit{ELSE normal} \rightarrow \textit{ACCEPT}$$
$$\textit{END IF}$$

Using such as methods instead of crisp/conventional logic (AND/OR/NOT) operation can reduce the false positive alarm rates in the system.

# CHAPTER 4

# Research Methodology

## 4.1 Research Overview

This study is divided into two major parts, first the theoretical part which involve on grasping an adequate in-depth study in the field of fuzzy logic (imprecision decision making perspective), linguistic based approach (rules based system or IF-THEN rules), and evolutionary algorithms, thus the qualitative methods will be used to getting more understanding of the subject and all pertinent disciplines. The second part of this study is an implementation (experimental) of how to integrate soft-computing methods with open-source software as prototyping to build self-adaptive and intelligent network protection system.

The first part of this study imposed to be as literature survey, which will use the qualitative methods to:

1. Research problem to be delineated precisely.

2. Label and identify work variables.

3. Produce hypotheses. (Sekaran, 2003)

The latter part of this study consists of design the prototype, which a software based system using open-source technology such as Linux[1], kernel based Netfilter modules with application layer patch, and the an open-source IDS (i.e Snort/Bro) [2] [3]. According to the software website; snort is the most widely deployed intrusion prevention technology in the world[4].

Consequently, in this phase (prototyping) an experimental research will apply, since we will need to test the hypothesis (can we build hybrid, intelligent self-adaptive system!) provided through some types of experiment such as fuzzy based algorithms combined with genetic and/or biological-inspired techniques. In this case some statistics methods may apply during the experimental phase, such as collecting raw data from network activity, analyse it and check how the current conventional security techniques (firewall and IDS) handle it, comparing the analysis findings with our proposed hypothesis and or prototyping. Using experimental methods in the second part to:

1. Test the proposed fuzzy-evolutionary based algorithms using described testing steps shown in appendix B.

2. Evaluate the output of collected data to the new proposed techniques.

3. Test performance and the final results.

4. Check the self-adaptively techniques proposed in the project.

---

[1] https://www.kernel.org/
[2] http://www.snort.org
[3] http://www.bro.org
[4] http://www.snort.org/snort

Therefore, a combine of descriptive and quantitative methods (for evaluation purpose), in addition to experimental method will be applied in this proposed project.

## 4.2  Research to be Undertaken

In this section the scope of work will be proposed; how the research objectives will be achieved? and what methods/tools will be used? first a profound understanding of the objectives and research focus is a mandatory task, as well as the pertinent literature review. Second phase, designing the algorithms that emerged from fuzzy-evolutionary approaches, followed by prototyping those algorithms to build hybrid intelligent system using the open-source tools.

### 4.2.1  Scope of Work

Why the use of open-source tools? simply because you have the source, which gives immense advantage of view the code, alter it and re-produce it as open-source, so other developers, researchers, and/or hobbyists view/alter and/or contribute with your research and give a positive feedback to the author(s), which indeed result enhancements and improvements towards a robust program and/or research.

> **The research method is based on TCP/IP protocol analysis in general and its headers in particular; and focus on one threat type [DoS], while we hope to be (the research idea, and aim) suitable for most of protocols and prevent 0-day threats.**

Decode each TCP/IP packet header for particular attributes (i.e. TCP flags, header length, TTL, etc...), then passes these parameters to the next level for further analysis as shown in the next chapter.

### 4.2.2   Software and technologies used during the research

During the design and experimental phase of this proposed phase, we will be using the following open-source tools. *Note: All the following is Open-Source software licensed under GPL license.*[5]

1. Linux kernel on Fedora OS (custom kernel 3.9.6).

2. Netfilter *(kernel based module)* with application layer patch.

3. Netfilter with fuzzy module support (see Appendix A).

4. Netfilter with string match support ($>$ kernel 2.6.18)

5. Snort and/or Bro as IDS/IPS.

6. Scilab community edition *(with fuzzy logic & genetic algorithms toolkit)*.

7. Custom scripts/codes in Bash/Perl and/or C.

## 4.3   How it Works

To prove the concept of this research a test environment has been prepared based on virtual machines (VirtualBox), as depicted in the following figure.

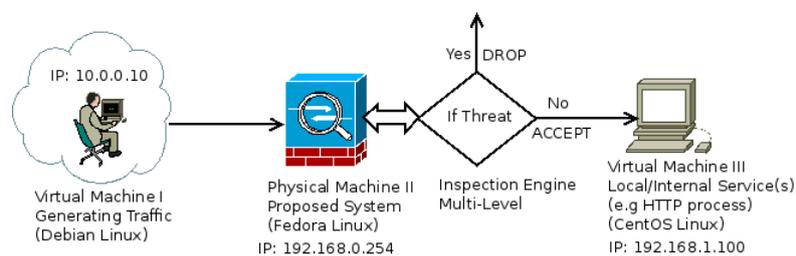

Figure 4.1: Working Environment

Two virtual machines were installed (on one physical machine) to prepare the test environment.

---

[5]http://www.gnu.org/licenses/gpl.html

Table 4.1: Test environment

| Machine | IP | Host Name | Purpose | Operating System |
| --- | --- | --- | --- | --- |
| Machine V-I | 10.0.0.10 | testvm3.local | Send malicious traffic | Debian Linux 7 |
| Machine P-II | 192.168.0.254 | test.local | Proposed system | Fedora Linux 18 |
| Machine V-III | 192.168.1.100 | testvm2.local | Internal server | CentOs Linux 6.4 |

V = Virtual Machine  P = Physical Machine

## Tools used

Tools used through the test environment in order to generate traffic for test purposes are:

### Machine I

- Default Linux *ping* command with -f option → flood ping with zero interval (see Appendix B).

- Use of hping3 [6] with –flood (see Appendix B)

- Port scanning using *nmap*[7] (see Appendix B)

- Using Apache HTTP server benchmarking tool (Linux *ab* command) (see Appendix B)

- Use of *nc* command (see below) to flood service.local machine with DoS attack.
  *seq 1 1000000 | nc -u service.local 4321* » -u = UDP » this command will send a million udp packets to machine III.

### Machine II

Proposed system with scripts to prove the idea.

- Fedora Linux with kernel 3.9

- IPtables version 1.4.19 with most recent extensions.

- IPset kernel module.

---

[6]http://www.hping.org/
[7]http://nmap.org/download.html

- Custom header sniffer script, Snort, Bro, and/or Suricata as packet sensors.
- Tcpdump as simple (but powerful) packet sniffing tool.
- Bash script to check sensors logs.
- Bash script using GA techniques to search/find best rule match.
- Fuzzy and recent iptables module to adapt fuzzy part.

**Machine III**

Running Apache web server (HTTP) with simple html files (just for testing purpose).
Another simple method to test, by using *nc* (Ncat) command:
Ncat *nc* is a feature-packed networking utility which reads and writes data across networks from the command line. Ncat was written for the Nmap Project and is the culmination of the currently splintered family of Netcat incarnations. It is designed to be a reliable back-end tool to instantly provide network connectivity to other applications and users. Ncat will not only work with IPv4 and IPv6 but provides the user with a virtually limitless number of potential uses.[8]

- Open new port and listen to the incoming connection:
  *nc -u -l 4321* » *-u* = UDP » *-l* = Listen to port 4321
  Now a network port has been opened using Ncat tools, when sender.cracker ran nc with packet flooding option, the first few packets from machine passed through, but then it stalled completely and no further traffic was received.

---

[8]Linux man pages

# CHAPTER 5

# System Architecture

Before going into the proposed system architecture, requirement/specification of the system should be described. System specifications summarised in the following table:

Table 5.1: Proposed system architecture

| Operating System | Fedora Linux with kernel 3.9.6. |
|---|---|
| Sensor | Custom script, kernel logs, tcpdump, Bro, and/or Snort. |
| Algorithms | Fuzzy-Genetic algorithms (generate rules). |
| Kernel Add-Ons (Patches) | Fuzzy and Recent patches. |
| System Logs | For learning mechanism. |

In this chapter a very short overview on TCP/IP protocol will be presented to understand the network packet flows, in order to grasp the necessary information of nodes communicate within a simple network environment; this simple overview should apply on any network regardless the scale.

## 5.1 Basic TCP/IP

Since this research study is concerning about network protection, a very brief overview about the TCP/IP basics will be helpful.

### 5.1.1 Inside the Packet

Packet is the main data structure on any network that communications between nodes encapsulates packet or sequences of packets. Each packet consists of a header, which gives valuable information about network routing, such where packet came from (source) and where going to (destination) plus other information (i.e. source port, destination port, flag, sequence number), and attached the main payload (data) into the tail of the packet, see figure (5.2).

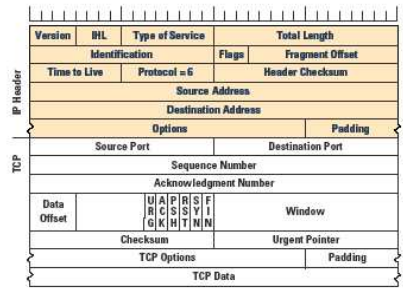

Figure 5.1: TCP/IP packet header structure

Image source: *http://www.bamed.org/2012/11/10/anatomy-of-a-packet/*

Network communications are broken into a series of relatively small packets, and sent out on the communication medium. During packets routing journey, its nominal dynamically routed. Each router determines the best next hop for the packet and sends it along to the next router closer to the destination. This means that different packets of the same communication can take different routes, and thus do not necessarily arrive in the order in which they are sent. ([Wegman and Marchette](), [2003]())

The main focus in this design phase on extracting some valuable TCP/IP header data to assist on classification phase, as our aim to make process simple as much as we can, hence it is easier to focus on packet headers instead of payload or profiling network behaviour.

**Why focusing on packet headers not its payload**, each network packet consists of header and payload (data), most of the Internet connections are using application layer (OSI layer 7)[1] protocols, such as HTTP, SMTP, DHCP, etc..., as a result most payload depends on application layer protocols and session environment. Each of these protocols has its own specification and diverse of payload size and signatures, which is dynamically changing accordingly.

Therefore, it will be hard (in the current stage) to investigate each protocol for deep packet inspection.

On the other hand, packet headers are constant and have many fields which can be formulated in order to discover abnormal packet behaviours, a previous endeavours was made ([Mahoney and Chan](), [2001]()) ([Wegman and Marchette](), [2003]()) to detect anomalous using packet headers, by using packet header analysis, which led to promising results.

---

[1] http://en.wikipedia.org/wiki/OSI_model

### 5.1.2 The Three-way Handshake

The three-way handshake is simply the source host and the destination host requesting a connection, and then confirming to each other that a connection has been made. To open a session a client determines a local source port and an Initial Sequence Number (ISN). The ISN is a randomly determined integer between 0 and 4,294,967,295. Communicating hosts exchange ISNs during connection initialization. Each host sets two counters sequence and acknowledgement. In the of a single TCP context packet, the sequence number is set by the sending host, and the acknowledgement number is set by the receiving host.

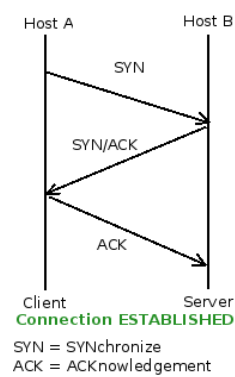

Figure 5.2: TCP Three-Way Handshake

Three-way handshake is significant and mandatory in Transmission Control Protocol (TCP) to establish a reliable connection, unlike User Datagram Protocol (UDP) that has no connection reliability. This process (3-way handshake) in TCP is needed between client and server to regulate initial sequence numbers (ISN) and ensure that they have both understood each other. The ISN should be unpredictable to defend against TCP sequence prediction attack (is an attempt to predict the sequence number used to identify the packets in a TCP connection, which can be used to counterfeit packets). [2]

### 5.1.3 A brief description of TCP flags

The TCP header contains several one-bit boolean fields known as flags used to influence the flow of data across a TCP connection. Unusual use of TCP flags is a good indicator of suspicious traffic.

---

[2] http://en.wikipedia.org/wiki/TCP_sequence_prediction_attack

Table 5.2: TCP flags

| Flag | Binary | Decimal | Meaning |
| --- | --- | --- | --- |
| URG | 00100000 | 32 | Packet segment is urgent and should be prioritized. |
| ACK | 00010000 | 16 | Acknowledges received data. |
| PSH | 00001000 | 8 | Push data immediately, rather than enter the buffer. |
| RST | 00000100 | 4 | Aborts a connection in response to an error. |
| SYN | 00000010 | 2 | Initiates a connection. |
| FIN | 00000001 | 1 | Finish/Close a connection. |
| SYNACK | 00010010 | 18 | Server send back confirmation to the client. |

Source: Admin Magazine [3]

TCP flags are significant for network behaviour, since most popular grid and Internet in particular are using TCP/IP protocols, hence a brief overview of one of the major parts in TCP header (flags) is mandatory. According to RFC793 [4] and other RFC's defined the way how systems should respond to packets, but they don't explain how systems should handle illegal combinations of flags.[5] At least one of six flags (SYN, ACK, RST, URG, PSH, FIN) must be set in each TCP packet; each flag corresponds to a particular bit in the TCP header.[5]

**A close look at TCP flags behaviour and how it is significant to network analysis,** this can be summarised in the following table.

Table 5.3: TCP flags behaviour explanation

| Flag Combination | Normal | Abnormal | Meaning |
| --- | --- | --- | --- |
| SYN, SYN/ACK, ACK | Yes | – | Required for 3-way handshake to establishes a TCP connection. |
| ACK | Yes | – | Every packet in a connection must have the ACK bit set. |
| PSH and/or URG | Yes | – | Optionally can be used after 3-way handshake for packet prioritisation. |
| FIN/ACK and ACK | Yes | – | Preparing to finish the connection. |
| RST/ACK or RST | Yes | – | Requesting immediate connection termination. |
| SYN/FIN | – | Yes | Considered to be a malicious. |
| SYN/FIN/PSH | – | Yes | Any combination with SYN/FIN (malicious). |
| FIN flag only | – | Yes | Used for port scan. |
| NULL flag | – | Yes | TCP flags cannot set to 0 (invalid packet). |

Furthermore, If any packets on top of the Internet Protocol (TCP, UDP, etc...) have a source or destination port set to 0 should be considered invalid packet, hence discarded, also it is important that for the acknowledgement number never be set to 0 when the ACK flag is set. A SYN only packet, which should only occur when a new connection is being initiated, should not contain any data. Packets should not use a destination address that is a broadcast address, usually IP ending in .255.[6] [7]

---

[3]http://www.admin-magazine.com/Articles/Intruder-Detection-with-tcpdump/28offset29/2
[4]http://www.ietf.org/rfc/rfc0793.txt?number=793
[5]http://www.symantec.com/connect/articles/abnormal-ip-packets

### 5.1.4 Data Collection from Single TCP Connection

- Protocol type → *(i.e. TCP, UDP, ICMP, etc ...)*

- Source → *Source IP, source port, and number of data bytes from src to dest.*

- Destination → *Destination IP, Destination port, and number of data bytes from dest to src.*

- Flag → (i.e. SYN, ACK, URG, PSH, FIN, RST)

- Packet duration → Length of connections (in seconds)

- Window size → Number of bytes can be allocated of buffer space.

- Urgent packet number.

- Network service on destination → *(i.e. FTP, HTTP, SMTP, etc ...)*

## 5.2 Attack Categories

According to KDDCUP'99 [8], network attacks fall into four main categories:

Table 5.4: Common network attack categories

| Category | Description |
| --- | --- |
| DoS/DDoS | Caused system unavailability (i.e. flooding) |
| R2L (Remote-to-Local) | Attempt to access from a remote machine (i.e. password guessing) |
| U2R (User-to-Root) | Attempt to access to local superuser (root) (buffer overflow) |
| Probing | Port scanning |

---

[6]http://packetcrafter.wordpress.com/2011/02/13/tcp-flags-hackers-playground/
[7]http://www.symantec.com/connect/articles/abnormal-ip-packets
[8]http://kdd.ics.uci.edu/databases/kddcup99/kddcup99.html

### 5.2.1 DoS/DDoS Attack

As described in section 1.9.2, a Denial of Service (DoS) attack usually involves attackers flooding the targeted system with massive amount fo regular or irregular to run out the system/service resources resulting system failure. While Distributed Denial of Service (DDoS) attack is a DoS attack utilizing multiple distributed attack sources.

### 5.2.2 Yet Another DoS Attack (Smurf Attack)

Smurf attack is targeted ICMP protocol by spoofing ICMP echo request (ping) packets, which are sent to a subnet broadcast address. This will cause each active host to send an echo reply to the source. In this attack the source address is set to the address of the target.

Since a large number of echo reply will be sent to the sender, which will causing degradation of service on its network. The Smurf attack exploits the concepts of packet amplification and address spoofing to overwhelm the target network. (Templeton and Levitt, 2003)

## 5.3 The Big Picture, but Still Simple

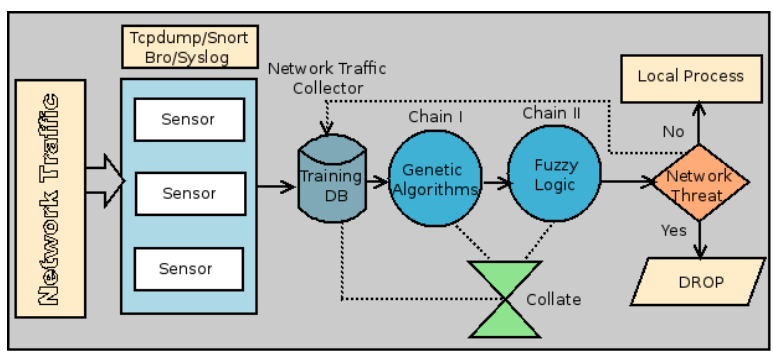

Figure 5.3: The Big Picture

### 5.3.1 Real-Time logging structure

Linux kernel (netfilter module) has a efficient capabilities to log network traffic (inbound and outbound) on every network interfaces and on real-time, therefore in this research the advantages of this capability has been taken in order to extract packet headers for processing phase. Another *c* script (appendix c) also has been used to extract packet headers (26 fields) logged by Linux kernel (networking sockets).

**Linux kernel netfilter logging explanation**

Output of Linux kernel netfilter/iptables looks like the following:

log-prefix: IN=eth0 OUT=eth0 MAC=00:00:00:00:00:00 SRC=192.168.50.1 DST=192.168.60.1 LEN=60 TOS=0x00 PREC=0x00 TTL=64 ID=79822 DF PROTO=TCP SPT=54380 DPT=80 SEQ=4207658428 ACK=0 WINDOW=43690 RES=0x00 SYN URGP=0 OPT (0204SSR40401450G01A739001010010002010214)

Table 5.5: Linux kernel network traffic log entries

| Log Entry | Description |
| --- | --- |
| IN | Input interface |
| OUT | Output interface |
| MAC | Ethernet hardware address (MAC address) |
| SRC | Source IP address |
| DST | Destination IP address |
| LEN | Packet length |
| TOS | Type of Service (for packet prioritization) |
| PREC | Precedent bits |
| TTL | Time to Live |
| ID | Packet identifier |
| PROTO | Network Protocol (i.e. TCP, UDP, ICMP) |
| SPT | Source port |
| DPT | Destination port |
| SEQ | Packet sequence number |
| ACK | Acknowledge bit set (Packet flags - i.e. SYN, ACK, FIN, etc ...) |
| WINDOW | Size of TCP window |
| RES | Reserved bits |
| URGP | Urgent packet |
| OPT | Options (Maximum Segment Size [MSS] Option Data) |

The other log file created by custom *c* scripts (listing on appendix c) providing additional information, which is missing in generic kernel log. The below figure shows more information that will be considered. Therefore a use of this custom script will be beneficial, more details about script output found in appendix c.

```
       PKTNR=5 Time=25-10-2013 21:18:30.332 IPV=4 IHL=20 TOS=0
        PKTSIZE=60  ID=9540  TTL=64  PROTO=6 CHKSUM=43788
 SRC=10.0.0.10 DST=192.168.1.100 SPT=50409 DPT=80 SEQ=3461283511
   ACKNR=0 TCPHL=40 URG=0 ACK=0 PSH=0 RST=0 SYN=1 FIN=0
       WIN=12600 TCPCHKSUM=31935 URGP=0  TCPIP_HDRL=60
```

Figure 5.4: Custom log output, generated from custom script

## 5.4 Design

The proposed system consists of 3 major components, first the sensor(s) (Snort, Bro, suricata, tshark, tcpdump, and/or custom scripts), in this stage (collection) the packet header data is substantial, thus the main point is to find a way to extract packet header information and sort it out; then pass it to the next level. Second is the processing engine, where the main function is to search for the intrusions and/or threats based on genetic algorithms. The third component is the filter engine that gives additional strength to the processing phase by applying fuzzy logic (as fuzzy classifier) if-then based rules resulting the final decision.

Using multi sensor needs further research, since the big picture of the proposed system would consists of multi sensors (Snort, Bro, Suricata) and cluster environment called cell-based cluster, which I would propose in further research, therefore in this research a single sensor will be applied (tshark, tcpdump and/or custom script, in addition to firewall log).

### 5.4.1 Overview

1. **Sensor**: Just, any network packet sniffer (tshark, tcpdump/snort/custom script and/or generic firewall log) → Simple script to extract packet header from the sniffer output then;→ Send it to log file (DB) for further processing.

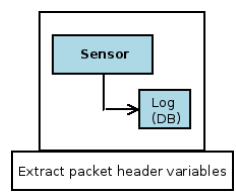

Figure 5.5: Sensor flow

2. **Processing Engine:** Genetic Algorithms (GAs) are general purpose optimisation algorithms and search method that mimic nature (biological) evo-lution and genetics to evolve a group of solutions to a problem. Figure (5.5) depict how generic GA works.

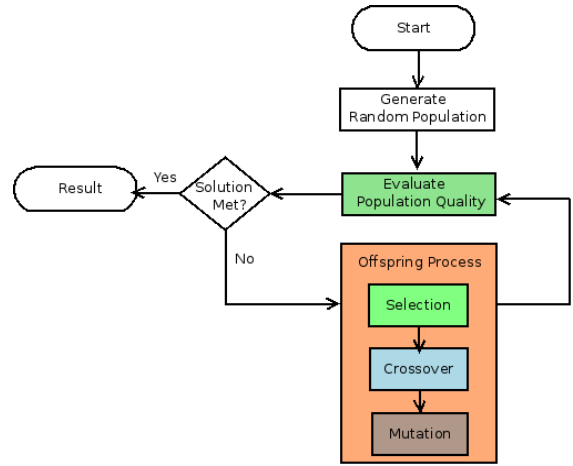

Figure 5.6: Generic GA flow chart

```
GA() {initialize random population;
evaluate population (fitness);
while termination conditions not met do
select solutions for next population (offspring);
perform crossover and/or mutation;
evaluate population (again fitness);
end while;
}
```

3. **Filter:** Preceding layer (Processing Engine) will send suspected ip including flag to filter engine (current stage) with a degree of values between 0 and 1 (i.e. 0.0, 0.1, 0.2,..., 1.0) based on suspicious packets header values and counts (network threat/intruder or not), in which fuzzy logic can determine the final decision, either allow the IP/packet of drop it.

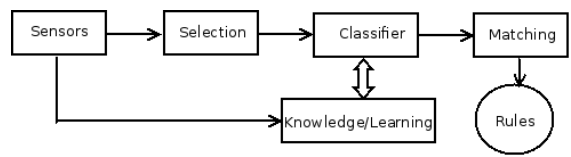

Figure 5.7: Rules classification

## 5.5 Genetic Algorithms obstacles

Evolutionary techniques is not a panacea, it has a certain limitations, however **it can be overcome**, whereas biological evolution is an undirected process of nature, which unfit element will be weeded out and who overcome the obstacles (fittest element) will continue. As all evolutionary algorithms (Genetic Algorithm one of most popular EAs) emulate nature process in general and biology in particular, likewise genetic algorithms obstacles can be overcome and none of them bear on the validity of biological evolution.[9]

Summarising Genetic Algorithms limitation and how can overcome it:

1. Defining a representation for the problem.
   (A.) Define individuals (Population) as lists of numbers - binary-valued, integer-valued, or real-valued - where each number represents some aspect of a candidate solution.
   (B.) Build individuals as executable trees of code that can be mutated by changing or swapping sub-trees.[9]

2. How to write the fitness function (If the fitness function is chosen poorly or defined imprecisely, the genetic algorithm may be unable to find a solution to the problem).
   The type and strength of selection - must be chosen with care.[9]

3. In the small populations premature convergence problem could be emerged (Chance variations in reproduction rate may cause one genotype to become dominant over others).
   To overcome this point, controlling the strength of selection, so as not to give excessively fit individuals too great of an advantage.[9]

It is not that genetic algorithms cannot find good solutions to such problems; it is merely that traditional analytic methods take much less time and computational effort than GAs and, unlike GAs, are usually mathematically guaranteed to deliver the one exact solution. Of course, since there is no such thing as a mathematically perfect solution to any problem of biological adaptation, this issue does not arise in nature.[9] (Holland, 1992) (Forrest, 1993)

## 5.6 Prerequisites & Encoding

Four steps needs (initially) to be performed, as a first preparation step (stage I), to feed the next layer (stage II - processing engine) with the required data

---

[9] http://www.talkorigins.org/faqs/genalg/genalg.html#limitations

representation (encoding data for GA), these steps are:

1. Collect data (packet header).
2. Filter data (match specific information).
3. Sort it out (data).
4. Encode Information - data representation (feed GA engines).

Data presented in decimal values for simplicity, also because most of extracted fields from network traffic (header) in decimals (i.e. port number, flags, etc...), except IP address (decimal dotted), which has been converted to decimal value (appendix c - section B)

## 5.7 Process (Step by Step)

There are many types of DoS attack (i.e. smurf, flood, null flood, etc...), but usually DoS involves a flood of packets, hence a particular header values will important to differentiate the DoS attack form normal packets, these header fields (more details in appendix c) are:

Table 5.6: Header fields to be used by process engine

| Field (18) | Description |
| --- | --- |
| Time | Timestamp |
| TOS | Type Of Service |
| PKTSIZE | Payload (data) size |
| TTL | Time To Live |
| PROTO | Protocol [1=ICMP, 2=IGMP, 6=TCP, 17=UDP] |
| SRC | Source IP |
| DST | Destination IP |
| SPT | Source Port |
| DPT | Destination Port |
| URG | Urgent flag |
| ACK | Acknowledgement flag |
| PSH | Push flag |
| RST | Reset flag |
| SYN | Synchronise flag |
| FIN | Finish flag |
| WIN | Packet Window size |
| URGP | Urgent flag Pointer |
| TCPIP_HDRL | TCP/IP Header Length |

2. In the first stage (layer) Sensor engine will count packets and extract the above fields and send it to log file (Leaning Database).

2. Threshold packets count will set to $> 500$ (count of packets received in particular time intervals), that why a need of extracting time-stamp and TTL fields are mandatory.
   But, why threshold set to be greater than 500 packets, because each operating system has a limited network stack table (receive buffer - physical memory allocation size), as this project using Linux kernel, which has a high performance network stack capabilities, the default Linux kernel receive buffer size is 2129920 bytes (2080KB). Therefore, assuming each initial network connection will consume $< 1KB$ (1024 Bytes), so default buffer size will handle $> 2000$ connection per cycle (this is just a minimum - Linux can handle more beyond this number), accordingly 500 packets is %25 percent of maximum buffer size, hence we choose this count as our threshold and consider any packets reach this limit for further analysis (go up to the next stage).
   More about Linux kernel network stack can be obtained by the following commands:
   *# sysctl -a | grep net | grep mem* $\rightarrow$ Check Linux kernel network stack buffer (of-course you can increase these values, depends on total memory size).
   *# ss -m* $\rightarrow$ show memory statistics used by kernel network stack.

3. Based on <u>fuzzy logic</u> (linguistic based) rules IF-THEN (in addition to the above), sensor engine will select which packet will go up to the next level.

$$\text{IF packets} > 500 \text{ AND time intervals HIGH THEN (threat} \rightarrow \text{DROP)}$$
$$\text{ELSE normal} \rightarrow \text{ACCEPT END IF}$$

4. <u>Genetic Algorithm</u> chromosomes represented in decimal values for the ease of use, one chromosome will be represented as decimal values combination {0,1,2,...,9} and semicolon (:) as separator.

Table 5.7: GA chromosomes representation by any decimal values combination

| 0 | 1 | 2 | 3 | 4 | 5 | 6 | 7 | 8 | 9 | : |
|---|---|---|---|---|---|---|---|---|---|---|

Table 5.8: Another example for chromosome combinations

| 1230 | : | 225 | : | 25 | 51770 | 4446 | : | 8080 | : | 1500 |
|------|---|-----|---|----|-------|------|---|------|---|------|

5. Chromosomes will be constant (only decimal values), but the search space (db) will dynamically changing, based on fuzzy rules and packet behaviour, chromosomes will be the target solution which will search in the learning database until matching, then generate rule(s).

6. Learning database (LDB) currently is a text based log file, which will be the search space for GA (process engine), the log file format will be as follow (very simple format):

    3232236056:25233:3232236328:80:1500:6

    Where semicolon (:) is fields separator, accordingly first field is source ip (in decimal) = 192.168.2.24, second field is source port 25233, third field is destination ip (in decimal) = 192.168.3.40, followed by destination port 80 and packet size 1500 bytes, then the last field is protocol 6 = TCP.
    The above format will be organized by the preceding layer (sensor and initial processing engine), also we can change this format dynamically based on packets behaviour (see further research - chapter 7, section 7.2).

7. The above format will be the GA target string, and it will be stored in different log files until the search engine match the desired criteria (described above).

8. If process engine successfully match the pattern, then it will create a firewall (iptables) rule and then will block or ban the suspicious traffic for a certain period of time, it is a automatic process (adaptive).

9. After rule has been generated, the system will record it in the learning database with some flags for future connection analysis, as a kind of tracking the packets behaviour, and as a one of learning methods (needs further research).

# CHAPTER 6

# Prototype and Evaluation

## 6.1 Prototype

In the current stage of this proposed system, as an early prototype, it consist of three major software engines using the available open-source tools (resources) that can assist to achieve project objectives. Much endeavours was made to run the proposed system in real-time mode, therefore a simple testing and evaluation environment (refer to chapter 4) has been implemented trying to achieve the evaluation criteria (next section). Eventually, the prototype consists of open-source based software running in top of Linux based *OS*, but our broader idea is to implement this project in hardware based appliance (further research), with more features and enhancement will be implemented; this prototype is the kernel of the our future work.

### 6.1.1 In Nutshell

As described in chapter 5 in more details, following is a synopsis recall from the preceding chapter:

– All is software based (user-space), there is no direct interaction to the kernel and/or hardware.
– In this stage (current project), we only focus on packet header behaviours.
– **Components:**

1. Sensor Engine (can be any packet sniffer), with a collector script to sort out packet headers.
2. Processing Engine, simple script to apply genetic algorithm for threat (DoS) search.

3. <u>Filter Engine</u> (final stage), fuzzy logic classifier as additional module to emphasis on final decision, based on linguistic based criteria (IF-THEN).

## 6.2 Evaluation Matrix

Table 6.1: Project Evaluation Matrix

| Weight Score (0→1) <br> Evaluation Criteria | **0** <br> N=Not Met | **0.5** <br> P=Partially Met | **1** <br> M=Met |
|---|---|---|---|
| System Adaptation | Cannot generate any rules. No LDB* sync. | Partially generate rules. No LDB sync. | Generate rules. LDB sync, but no correlation. |
| System Intelligence | Cannot match patterns. No LDB sync. | Partially pattern matching. No LDB sync. | Match given patterns. LDB sync. |
| Processing Engine Efficiency (GA/FL) | No matching. Run out of resources. | Partially pattern matching. | Pattern matching. Rule generated. |
| Threats Detection Rate | <=50% | >50%, <=80% | >80% |
| Overall Efficiency | Run out of resources. | Delay | Partially delay. |

*LDB = Learning Database (Knowledge Base).
**Note:** Currently learning database (LDB) format is text based log file(s).

To test the main script performance (process engine) that apply the genetic algorithm method, the program run for ten times and best matching time has been recorded and calculated to obtain performance average, details in section 6.5

## 6.3 Results

As part of this research project, is to experiment and evaluate the proposed system, hence in this chapter an experimental results will be presented. First, detection rate equation is needed.

Table 6.2: Detection rate formula.

| Formula | Meaning |
|---|---|
| $R = \frac{a}{n} * 100$ | Where $R$ is detection rate, $a$ is actual packet detected, and $n$ is total packets count. |

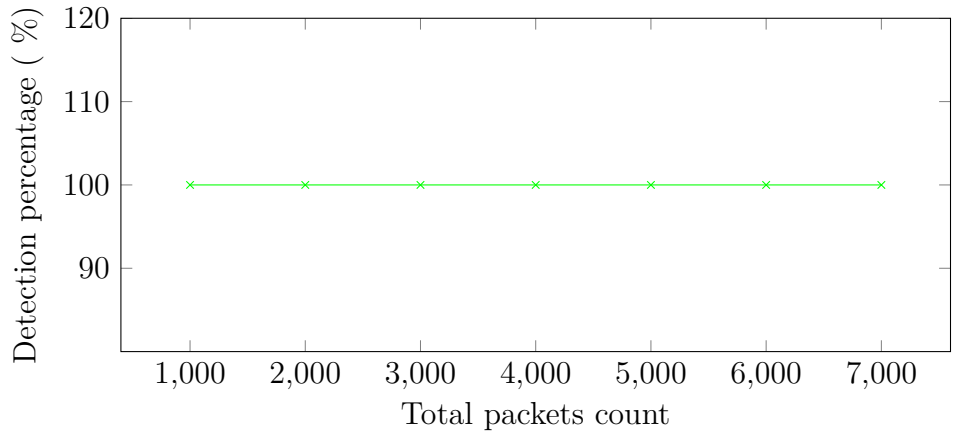

Figure 6.1: DoS threat detection rate (TCP)

In TCP packets (as depicted in the above figure) detection rate was high, sensor script counted the packets and filter it upon fuzzy rules (IF-THEN) and send it to db , then process script matching the pattern easily in average time $\approx$ 1.8 minutes (109 second).

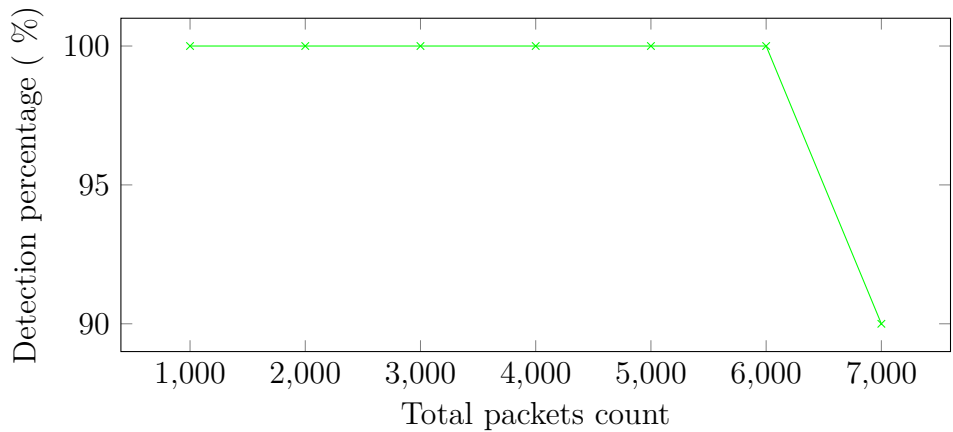

Figure 6.2: Ping flood threat detection rate (ICMP)

In ICMP ping flooding, the detection rate was dropped to %90 due to a very fast time intervals with high packets count, which bash script could not handle and cope with it, particularly if entries in the database more than 50.

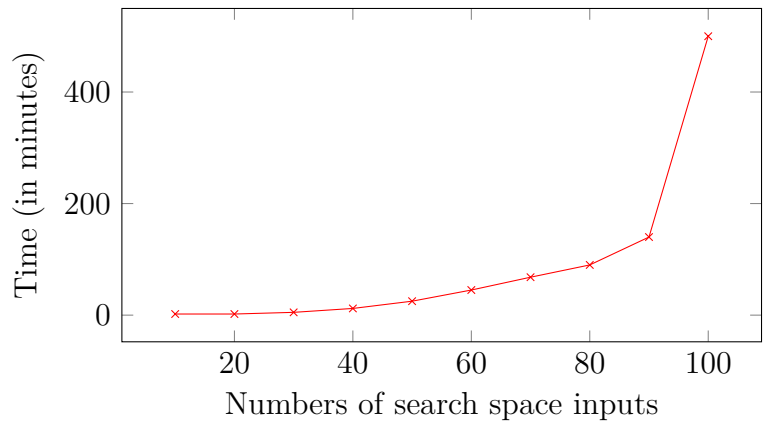

Figure 6.3: Search space delay based on log (db) entries

In figure 6.3 the program run out of resource (generation, time, and/or cpu utilization) when log inputs (database) counts is high, due to resource limitation as described in more details in section 6.5

## 6.4 Evaluation

In order to evaluate the current experimental results an evaluation matrix has been presented in section 6.2, just to make the evaluation as simple as it should be, therefore a straightforward table presenting degree of achievement of the current status.

Based on results and findings during the experimental phase, which has been presented in the previous section (6.3), the following table shows how far this results from the desired objectives.

Summering the project objectives and research focus:

- Study soft-computing (Genetic Algorithms & Fuzzy Logic) methods and apply it in network-based threat prevention system.
- Build adaptive system, automatic protection in real-time network traffic.

- Focus on one threat (DoS), but the concept should apply to many threats as well (further research).
- Using open-source tools, as major components of the proposed system.

Table 6.3: Current experimental evaluation

| Weight Score (0→1) / Evaluation Criteria | 0 N=Not Met | 0.5 P=Partially Met | 1 M=Met |
|---|---|---|---|
| System Adaptation | – | – | M |
| System Intelligence | – | – | M |
| Processing Engine Efficiency (GA/FL) | – | – | M with PD |
| Threats Detection Rate | – | – | M (TCP) |
| Overall Efficiency | – | P | – |

PD = Partial Delay

According to the experimental phase, the major observation is time delay issue, which described in the following section in more details, another thing to be considered is detection rate, particularly in smurf threat (see section 5.2.2), since this kind of attack using broadcasting IP's with very fast small intervals, which will try to flood your system with tremendous amount of unwanted packets, accordingly system log files (database) cannot log all IP's and check it whether broadcast IP or no, but this can overcome easily with performance enhancement and improvements in database structure. (refer to the next section) In spite of the above obstacle, the system indeed detect this attack based on packet intervals and constant payload size as described in chapter 5 (section 5.7), then count the packets and send it to the second layer (process engine), but due to the above description system cannot generate rules and/or block it.

In table 6.3 processing engine (GA) efficiency, where TCP/ICMP pattern matched and blocking rules have been generated for both tcp and icmp flooding, but we indicate PD (Partially Delay), due to the time delay in detecting icmp flooding.

Further, in table 6.3 detection rate row, according to experimental results the proposed system detect both tcp and icmp dos flood, but with icmp detection slightly lower than tcp due to fast time intervals as described in section 6.3 (figure 6.2), since tcp detection has been met the evaluation criteria in the maximum rate ( %100), we consider it as fully met.

## 6.5 Current Limitations

During the experimental phase some obstacles and limitations emerged, such as time efficiency in general and in processing engine in particular, due to scripting language (Bash) used and hardware resources, since there is a platform layer between user space and kernel space (the shell), as well as scripting language limitation, which caused time delay, while searching for optimum solution (threat(s) pattern matching).

Another obstacle, the database (log file) size, since the sensor script capture real-time network traffic, thus the log file is increased significantly (sometimes reached more than 300MB), accordingly parsing this file to get the desired output is extremely difficult and caused significant time delay.

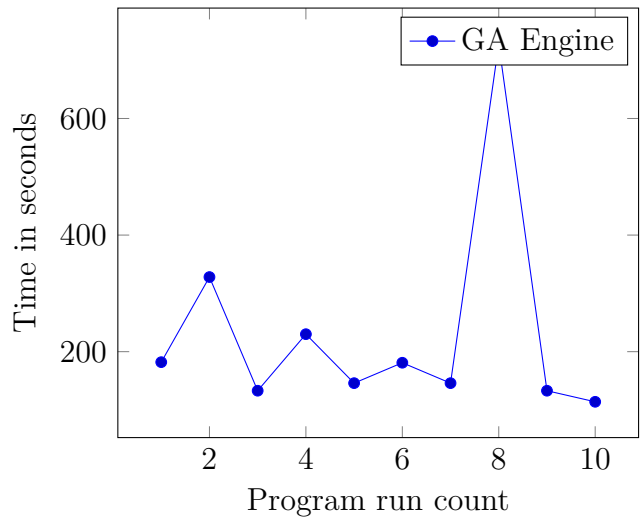

Figure 6.4: Process engine performance

$$\text{Performance mean point} = \frac{\sum \text{Time in seconds}}{\text{Program run count}}$$

$$\text{Mean} = \frac{182 + 328 + 133 + 230 + 146 + 181 + 146 + 728 + 133 + 114}{10}$$

$$\text{Performance avaerage: } \frac{2321}{10} = 232.1/60 = 3.8 Minutes$$

Furthermore, the chromosomes representation and search space is a major part in genetic algorithm, most of previous pertinent researches (chapter 2), have been represented chromosomes including genes as solutions candidates or individuals either in binary strings or decimal notation (Gomez and Dasgupta, 2002), (Li, 2004), (Gong et al., 2005); which theses values (chromosomes/solutions) needs to be known prior In this research project the chromosomes was constant, which consists of our generic genes (in decimal values), and the search space will evolve as described in chapter 5 (section 5.7), the consideration is, while we are in the testing phase, we observed the following issue: when the search space dynamically changed it caused a delay particularly if it has multi input, which have been added form previous system layer (sensor), and we believe this is due to limited resource as described above.

In addition, during the data process by GA engine we observed, if any error arise while the script generating the firewall rules, it will stop, hence will not continue to build up the rules. Therefore a need of genetic programming techniques is mandatory, which the system will be able to detect the error and evolve a solution to fix this error, this really need more investigation and further research; as our broader aim is build a total self-organized and intelligent system. Genetic programming is a method for getting computers to automatically solve problems and to evolve new generation of programmes that can be better than the preceding one.(Poli et al., 2008)

Most of this obstacles is performance based limitations, which can be overcome with more enhancement and tuning, also for large network scale a cell based (cluster) should be considered. Therefore, to overcome this limitation the code (processing engine script) could be changed to high-level compiled language such as $C$ in addition to log file rotation with copy concatenation to keep network traffic in real-time mode; also cell based (cluster) environment can be used for performance enhancement.

Further, a lightweight based agent (host/client based) could be implemented to collect network traffic behaviours based on particular pattern and send it to centralized server for learning process and data correlation analysing (further research - chapter 7).

# CHAPTER 7

# Conclusions and Further Work

## 7.1 Summary of Work

In this research project, simplicity was an aim to synthesise theories, methods, and tools to build self-organized intelligent network protection system with focus on combination of firewall and Intrusion Detection/Prevention mechanism, barely few research focused on such combination. Trying to design/develop flexible and adaptive security oriented approaches is a challenge, therefore in this research project an endeavour was made to design a novel approach for network threats protection system.

We studied fuzzy logic, genetic algorithms, artificial immune system, and major network threats, and we selected one of the most important network threat (Denial of Service), in order to emphasize about the above idea and to present a simple proof of concept, which is the ability to build an All-in-One (state of the art) intelligent network protection system; also, we have asserted that packet header is a major part to formulate packet behaviour, in addition to its simplicity.

**Our major endeavours of this dissertation can be summarised as follow:**

- Find novel approach for network security, by design a self-adaptive system that can be evolve within its operating environment.
- Design a fuzzy-genetic based classifier that can intelligently detect threats and then act, either to block the threat source or ban it for a certain period of time (this is very flexible), the main points is detection.
- Most of network security gateways are trapped in false positive/negative issues, which is means either allow connection that should not be allowed (threat) or the opposite, deny (block) normal connection and consider it as threat. Therefore we put this issue into our consideration,

   accordingly we tried to make simple design for packet header inspection, of-course it (still) needs more study and investigation, but the main concept is the same.

– Combine firewall and IDS techniques (All-in-One network threats protection system). Almost all current researches focused only either IDS (most) or firewall (few).

– Design simple GA mechanism to implement an automated intelligent system with no or slightly human interaction, our aim (future research) is to build a total adaptive and intelligent system (no human management at all).

## 7.2  Problems Encountered and Considerations

During this research project we have had some obstacles (details in section 6.5), which are a performance based issues, absolutely does not influence either the idea (project concept) and/or the overall promising results, but still needs further research to tackle all this obstacles. Therefore we can assert that the proposed concept (system) is durable and robust, of-course needs improvements and enhancement, which can be done by doing further research (see next section). The proposed system has been implemented using a laptop, which has limited resource (i.e. memory, HD i/o, etc...), in addition to virtual machines on top of the same hardware, hence to achieve a better performance an adequate resources is needed.

One of major obstacle the database (search space) scalability and db size (we used a text based files) particularly when having multiple inputs it cause delay for the output, that is due to (described in section 6.5) db size and inputs, in addition to bash script performance, which indeed needs to be converted to $C$ language.

## 7.3  Conclusions and Recommendations

We believe in integration between academic research and industry, hence an endeavours was made during this project trying combine theory with software based prototype. As a proof of concept this research focused on how to integrate unconventional methods (EA and FL) with open-source tools to build a simple intelligent and adaptive system that can detect threats either known (DoS) or unknown (further research - see next section).

Within this research project we tried to prove (as proof of concept) that a self-organized and intelligent system can be made with simplicity, using current available open-source tools combined with unconventional methods (soft-computing) to tackle one of the most significant in modern information environment, which is network threats.

Furthermore, in this research project we presents another step in detecting computer network threats through the use of a novel methods, which not only detect threats, but learn, choose and then act (adaptation), upon packet header behaviour.

A final word, we can assert, that both proposed idea and experimental system are promising, as well as durable, hence we will focus our most endeavour (further work) to make more improvements, enhancements, and adding more features and/or components to achieve the broader aim (state-of-the-art system).

**Recommendations**

- Build a packet behaviour schema (profile).
- All codes/scripts (programmes) presented in this research project should be converted into more efficient high-level compiled language (i.e. *C* or *C++*).
- GA search space should be enhanced (i.e resource limitation, such as text based files with multiple input).
- Learning database should be converted from text based files to indexing database for scalability.
- Implement genetic programming techniques (not genetic algorithms) as part of adaptation process.

## 7.4 Further work

The big picture of this research, is to build an All-on-One and state of the art intelligent, self-organized system that can be able to communicate with other devices, agent, terminals, and systems to detect either host-based or network threats whether known or 0-day attack. As a research will never stops, our endeavour to go for further research can be summarized as follow:

- Build a new network threats (intrusion detection) data set, since the most popular one [1] is very old (1998-1999).

  The new data set, will be very valuable for researchers and research development world wide in the area of networking threats detection, our aim is to include host-based threats signatures as well.
  <u>This will be very valuable add-on for the University</u>.

- Add host-based threats protection to the network-based.
- Using multi detectors (sensors) via distributed environment.
- Using multi agent (data collector) via distributed grid.
- Add scalability feature (distributed based system).
- Build a self-organized system (no human interaction at all), a rational system that can learn and then act.
- Add more improvements, tuning, and enhancement to the current idea.
- A complete security protection system, using cell-based cluster concept and biological inspired computation, as well as fuzzy based methods, with emphasis on the rise of the Internet of things concept.
- Build the proposed system in a lightweight appliance (software and hardware).
- Using a new concept of what called a cell-based cluster, which is an imitation of the simple living cell and how can communicate and interact with other cells in a beautiful harmony to achieve their tasks.

---

[1] http://www.ll.mit.edu/mission/communications/cyber/CSTcorpora/ideval/data/

# Appendix A

## A  Screenshots

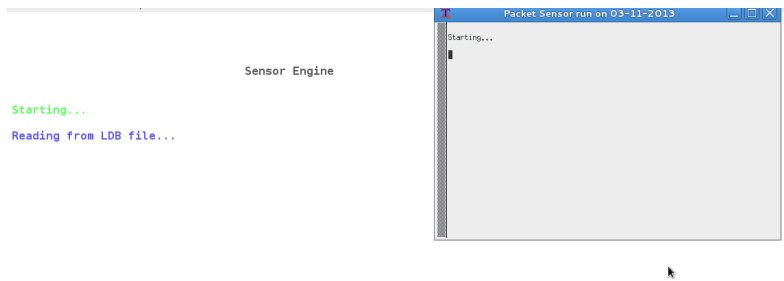

Figure 1: Sensor engine start header

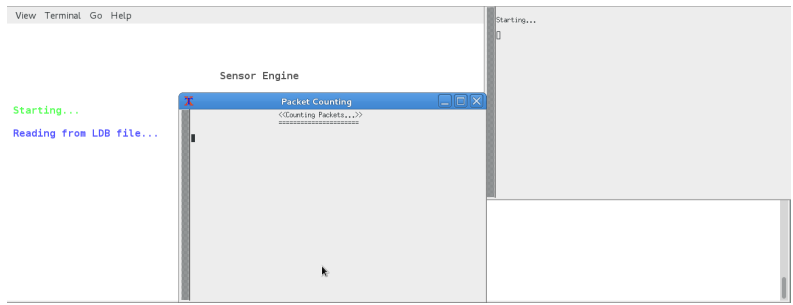

Figure 2: Sensor engine start header cont.

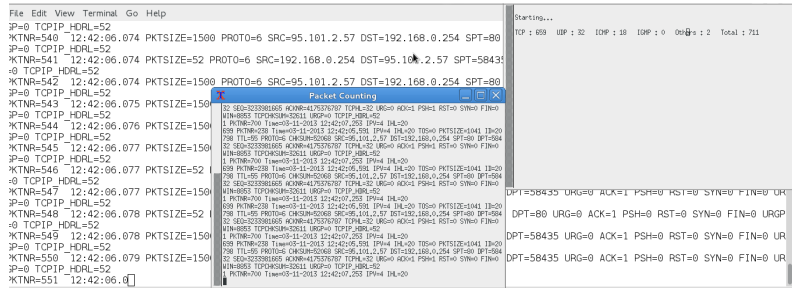

Figure 3: Sensor engine output

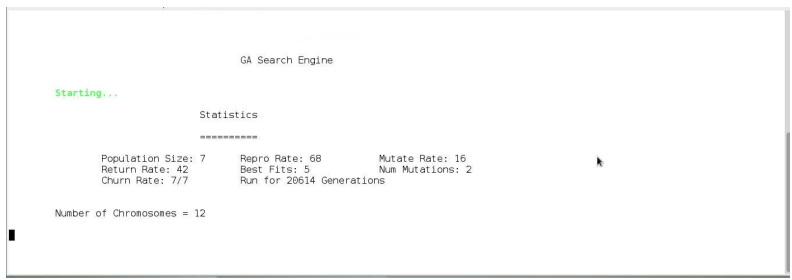

Figure 4: Process engine header

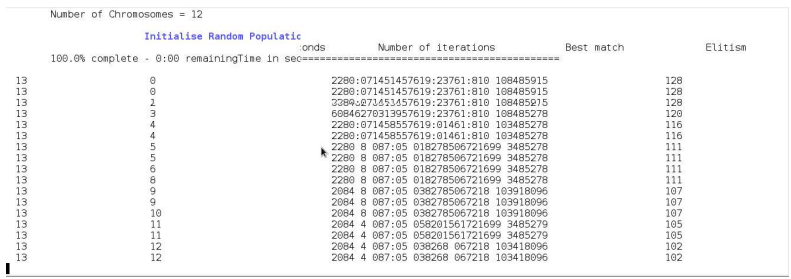

Figure 5: Process engine (GA) search display

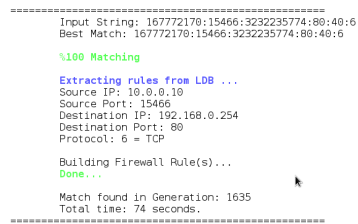

Figure 6: Process engine (GA) output

```
==================================================
        Input String: 167772170:3325:3232235774:80:160:6
        Best Match:   167772170:3325:3232235774:80:160:6

        %100 Matching

        Extracting rules from LDB ...
        Source IP: 10.0.0.10
        Source Port: 3325
        Destination IP: 192.168.0.254
        Destination Port: 80
        Protocol: 6 = TCP

        Checking Learning Database...
        Existing.. (No need to add more rules)

        Match found in Generation: 3133
        Total time: 109 seconds.
==================================================
```

Figure 7: Process engine while checking entries in learning database